\documentclass[10pt,twocolumn,letterpaper]{article}

\usepackage{3dv}
\usepackage{times}
\usepackage{epsfig}
\usepackage{graphicx}
\usepackage{amsmath}
\usepackage{amssymb}

\usepackage[ruled,vlined]{algorithm2e}
\SetAlFnt{\small}
\usepackage{siunitx}
\usepackage{booktabs}
\usepackage{makecell}
\usepackage{enumitem} %
\usepackage{adjustbox}
\usepackage{subfig}

\usepackage{array}
\newcolumntype{P}[1]{>{\centering\arraybackslash}p{#1}}
\newcolumntype{M}[1]{>{\centering\arraybackslash}m{#1}}

\usepackage{cite}
\usepackage{url}
\usepackage{titling}
\date{}\predate{}\postdate{}
\pretitle{\begin{center}\Large \bf}
\posttitle{\end{center}\vskip 0.5em}

\usepackage{color}

\threedvfinalcopy %

\def\threedvPaperID{245} %

\newif\ifshowpagenumbers
\showpagenumberstrue

\newcommand{\Fig}{Fig.~}

\newcommand{\up}[1]{{\color{black}#1}}

\newcommand{\mm}[1]{{\color{black}#1}}

\ifthreedvfinal
\ifshowpagenumbers
\else
\fi
\fi

\def\MYTITLE{ESL: Event-based Structured Light}
\title{\MYTITLE}

\newcommand\MYhyperrefoptions{bookmarks=true,bookmarksnumbered=true,
pdfpagemode={UseOutlines},plainpages=false,pdfpagelabels=true,
colorlinks=true,breaklinks=true,
pdftitle={\MYTITLE},%
pdfsubject={Robotics, Computer Vision},%
pdfauthor={M. Muglikar, G. Gallego, D. Scaramuzza},%
pdfkeywords={Depth sensing, Event camera, Asynchronous sensor, laser projector}}%
\usepackage[\MYhyperrefoptions,pdftex]{hyperref}

\author{Manasi Muglikar \textsuperscript{1}\\
\and
Guillermo Gallego \textsuperscript{2}\\
\and
Davide Scaramuzza \textsuperscript{1}\\
}
\newcommand\nomarkerfootnote[1]{%
  \begingroup
  \renewcommand\thefootnote{}\footnote{#1}%
  \addtocounter{footnote}{-1}%
  \endgroup
}

\definecolor{somegray}{rgb}{0.5, 0.5, 0.5}
\newcommand{\darkgrayed}[1]{\textcolor{somegray}{#1}}
\makeatletter
\newcommand*\titleheader[1]{\gdef\@titleheader{#1}}
\AtBeginDocument{%
  \let\st@red@title\@title
  \def\@title{%
    \vskip-4em
    \bgroup\normalfont\large\centering\@titleheader\par\egroup
    \vskip1.5em\st@red@title}
}
\makeatother

\titleheader{\darkgrayed{This paper has been accepted for publication at the\\
International Conference on 3D Vision (3DV), Online, 2021.
\copyright IEEE}}

\begin{document}

\maketitle

\nomarkerfootnote{\up{\textsuperscript{1}Dept.~Informatics, University of Zurich and Dept.~Neuroinformatics, University of Zurich and ETH Zurich, Switzerland. \textsuperscript{2}Technische Universit\"at Berlin and Einstein Center Digital Future, Berlin, Germany.\\
		This research was supported by SONY R\&D Center Europe and the National Centre of Competence in Research (NCCR) Robotics, through the Swiss National Science Foundation.
}}
\ifshowpagenumbers
\else
\thispagestyle{empty}
\fi

\global\long\def\bx{\mathbf{x}}
\global\long\def\TE{T_{pc}} %

\global\long\def\taup{\tau_{p}}
\global\long\def\tauc{\tau_{c}}
\global\long\def\bxp{\bx_{p}}
\global\long\def\bxc{\bx_{c}}

\begin{abstract}
Event cameras are bio-inspired sensors providing significant advantages over standard cameras such as low latency, high temporal resolution, and high dynamic range.
We propose a novel structured-light system using an event camera to tackle the problem of accurate and high-speed depth sensing.
Our setup consists of an event camera and a laser-point projector that uniformly illuminates the scene in a raster scanning pattern during 16 ms.
Previous methods match events independently of each other, and so they deliver noisy depth estimates at high scanning speeds in the presence of signal latency and jitter.
In contrast, we optimize an energy function designed to exploit event correlations, called spatio-temporal consistency.
The resulting method is robust to event jitter and therefore performs better at higher scanning speeds. 
Experiments demonstrate that our method can deal with high-speed motion and outperform state-of-the-art 3D reconstruction methods based on event cameras, reducing the RMSE by 83\% on average, for the same acquisition time. 
Code and dataset are available at \url{http://rpg.ifi.uzh.ch/esl/}.
\end{abstract}
\section{Introduction}
\label{sec:intro}
\global\long\def\figHeight{2.84cm}
\begin{figure}[t]
\centering
\subfloat[3D Scene]{\label{fig:eyecatch:photo}\includegraphics[trim={16cm 2cm 12cm 0},clip,height=\figHeight]{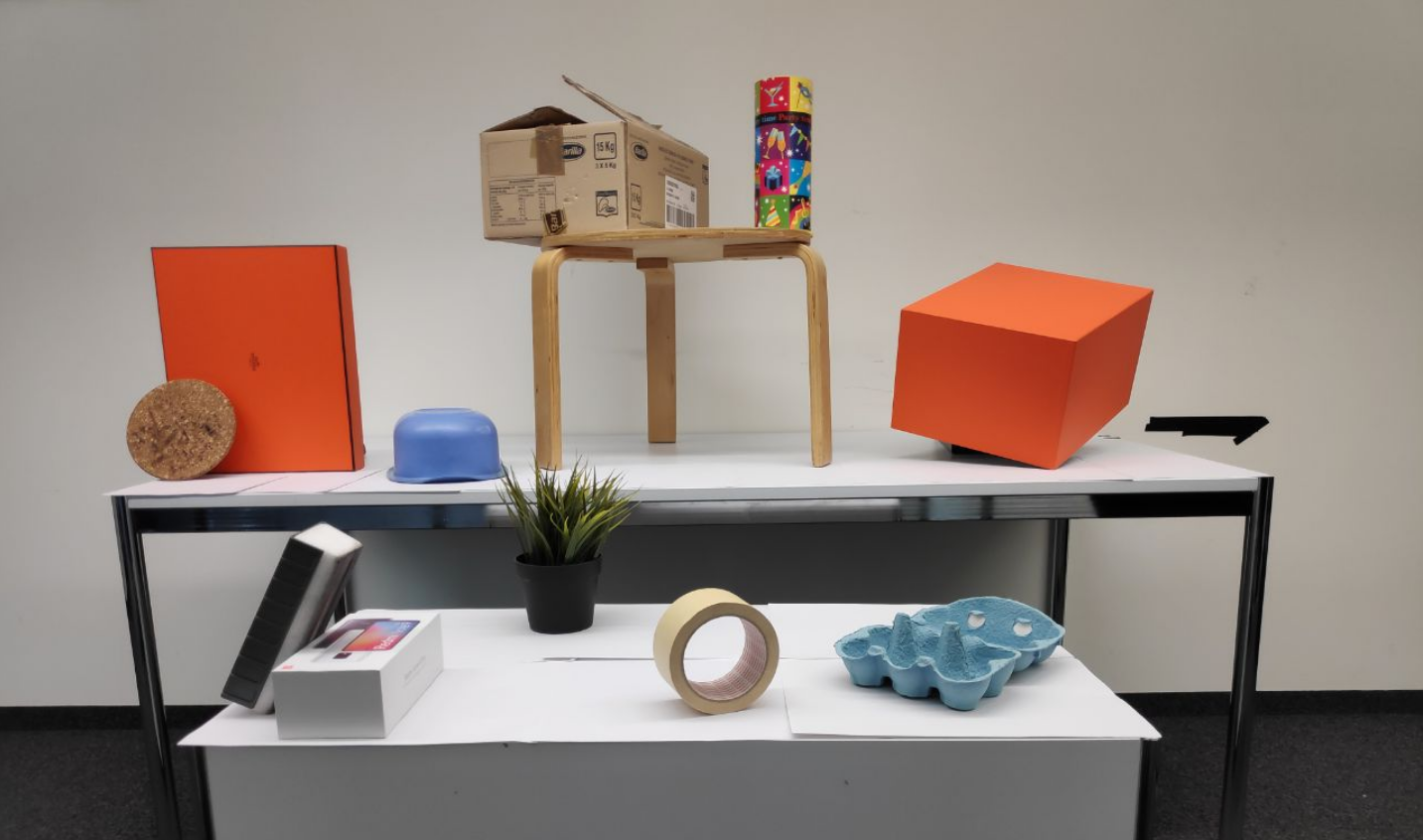}}\;
\subfloat[EMVS~\cite{Rebecq18ijcv}]{\label{fig:eyecatch:patch}{\includegraphics[trim={10cm 0 6cm 0},clip,,height=\figHeight]{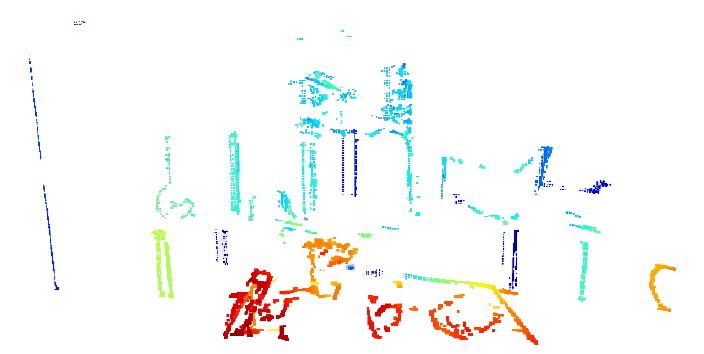}}}\;
\subfloat[Intel R.]{\label{fig:eyecatch:realsense}{\includegraphics[trim={19cm 10cm 22cm 9cm},clip,height=\figHeight]{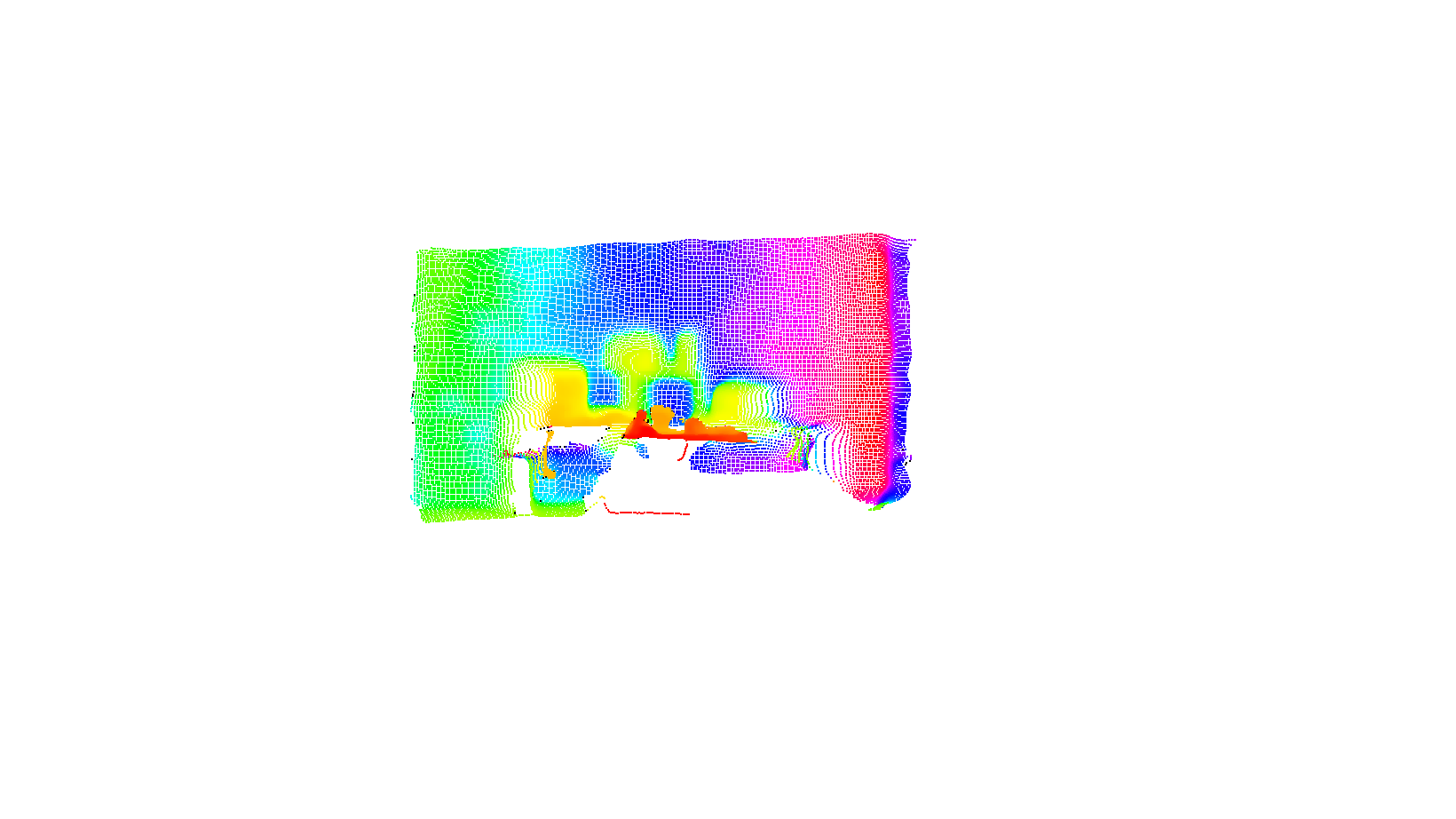}}}\;
\subfloat[Ours]{\label{fig:eyecatch:ours}{\includegraphics[trim={18cm 3cm 18cm 5cm},clip,height=\figHeight]{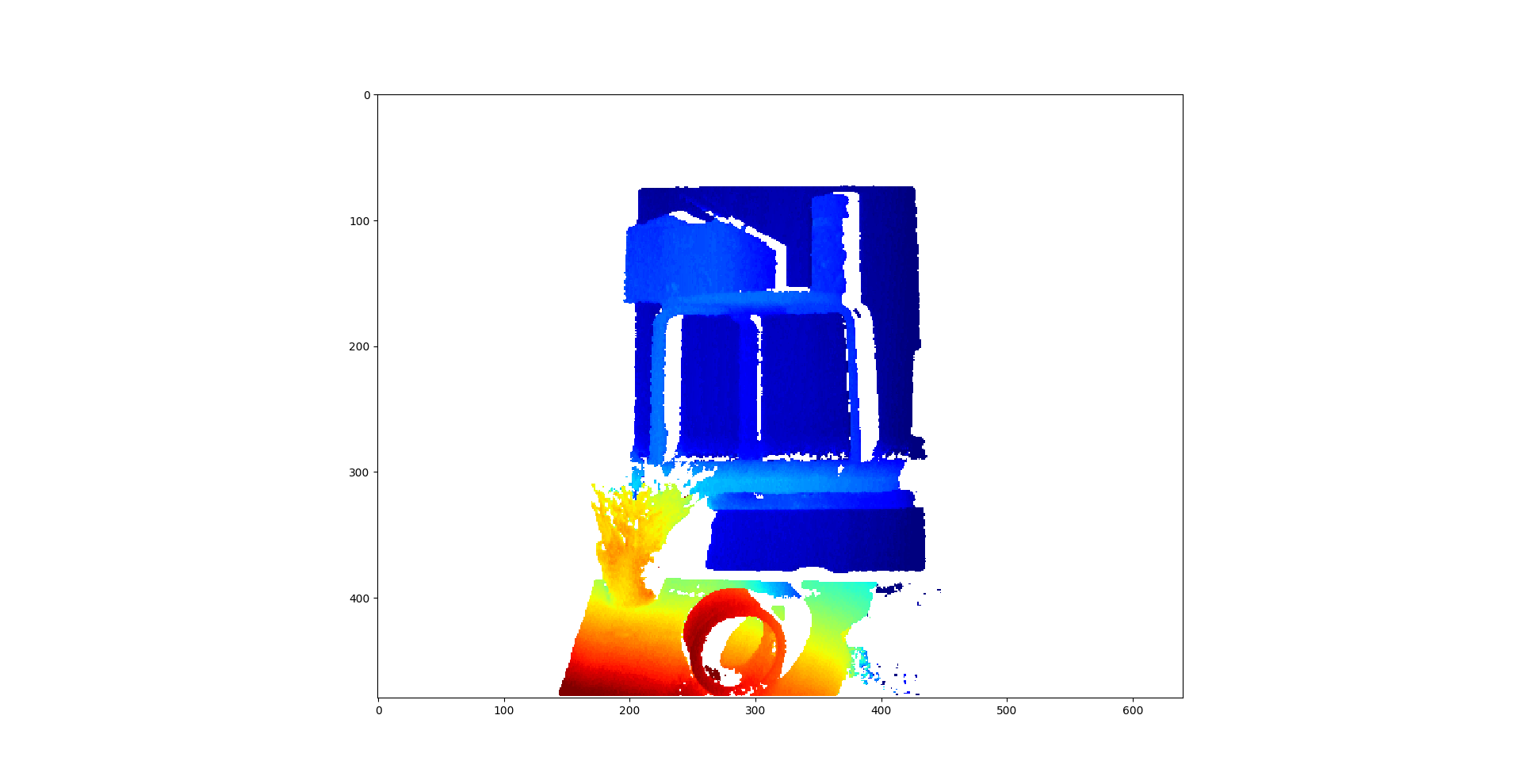}}}
\vspace{-1.5ex}
\caption{\emph{Depth estimation with a laser point projector and an event camera}.
    The proposed event-based method (d) produces more dense and accurate depth estimates than 
    (b) EMVS (events alone, without projector) or (c) a frame-based depth sensor (Intel RealSense D435).
    The color image (a) is not used; shown only for visualization purposes. %
    }
\vspace{-2ex}
\label{fig:eyecatch}
\end{figure}

Depth estimation plays an important role in many computer vision and robotics applications, such as 3D modeling, augmented reality, navigation, or industrial inspection. 
Structured light (SL) systems estimate depth by actively projecting a known pattern on the scene 
and observing with a camera how light interacts (i.e., deforms and reflects) with the surfaces of the objects.
In close range, these systems provide more accurate depth estimates than passive stereo methods, 
and so they have been used in commercial products like KinectV1~\cite{KinectV1} and Intel RealSense~\cite{IntelRealSense}.
Due to simple hardware and accurate depth estimates, SL systems are suitable for applications like 3D modeling, augmented reality, and indoor-autonomous navigation.

SL systems are constrained by the bandwidth of the devices (projector and camera) and the power of the projector light source.
These constraints limit the acquisition speed, depth resolution, and performance in ambient illumination of the SL system.
The main drawback of SL systems based on traditional cameras is that frame rate and redundant data acquisition limit processing to tens of Hz.
By suppressing redundant data acquisition, processing can be accelerated and made more lightweight.
This is a fundamental idea of recent systems based on event cameras, such as~\cite{Matsuda15iccp,Brandli13fns,Martel18iscas}.

Event cameras, such as the Dynamic Vision Sensor (DVS) \cite{Lichtsteiner08ssc,Suh20iscas,Finateu20isscc} or the ATIS sensor \cite{Posch11ssc,Posch14ieee}, are bio-inspired sensors that measure per-pixel intensity \emph{changes} (i.e., temporal contrast) asynchronously, at the time they occur.
Thus, event cameras excel at suppressing temporal redundancy of the acquired visual signal, and they do so at the circuit level, thus consuming very little power.
Moreover, event cameras have a very high temporal resolution (in the order of microseconds), which is orders of magnitude higher than that of traditional (frame-based) cameras, so they allow us to acquire visual signals at very high speed.
This is another key idea of SL systems based on event cameras: the high temporal resolution simplifies data association by sequentially exposing the scene, one point~\cite{Matsuda15iccp} or one line~\cite{Brandli13fns} at a time.
However, the unconventional output of event cameras (a stream of asynchronous per-pixel intensity changes, called ``events'', instead of a synchronous sequence of images) requires the design of novel computer vision methods ~\cite{Gallego20pami,Benosman14tnnls,Kim14bmvc,Zhu17icra,Rebecq17ral,Gallego17pami,Osswald17srep,Mueggler18tro,Gallego19cvpr}.
Additionally, event cameras have a very high dynamic range (HDR) ($>$\SI{120}{\decibel}), which allows them to operate in broad illumination conditions~\cite{Rosinol18ral,Rebecq19pami,Cohen18amos}.

This paper tackles the problem of depth estimation using a SL system comprising a laser point-projector and an event camera (Figs.~\ref{fig:eyecatch} and \ref{fig:setup}).
Our goal is to exploit the advantages of event cameras in terms of data redundancy suppression, large bandwidth (i.e., high temporal resolution) and HDR.
Early work~\cite{Matsuda15iccp} showed the potential of these types of systems; however, 3D points were estimated independently from each other, resulting in noisy 3D reconstructions.
Instead, we propose to exploit the regularity of the surfaces in the world to obtain more accurate and less noisy 3D reconstructions. 
To this end, events are no longer processed independently, but jointly and following a forward projection model rather than the classical depth-estimation approach (stereo matching plus triangulation by back-projection).

\textbf{Contributions}. In summary, our contributions are:
\begin{itemize}[topsep=1pt,parsep=2pt,partopsep=2pt]
\setlength\itemsep{0em}
    \item A novel formulation for depth estimation from an event-based SL system comprising a laser point-projector and an event camera.
    \mm{We model the laser point projector as an ``inverse'' event camera 
    and estimate depth by maximizing the spatio-temporal consistency between the projector's and the event camera's data, when interpreted as a stereo system.}
    \item The proposed method is robust to noise in the event timestamps (e.g., jitter, latency, BurstAER) as compared to the state-of-the-art \cite{Matsuda15iccp}.
    \item A convincing evaluation of the accuracy of our method using \mm{ten} stationary scenes and 
    a demonstration of the capabilities of our setup to scan \mm{eight} sequences with high-speed motion.
    \item A dataset comprising all static and dynamic scenes recorded with our setup, \mm{and source code}. 
    To the best of our knowledge it is the first public dataset of its kind.
\end{itemize}

The following sections review the related work (Section~\ref{sec:related-work}), 
present our approach \mm{ESL} (Section~\ref{sec:methodolody}),
and evaluate the method, comparing against the state-of-the-art and against ground truth data (Section~\ref{sec:experim}).

\section{Event-based Structured Light Systems}
\label{sec:related-work}
Prior structured light (SL) systems that have addressed the problem of depth estimation with event cameras are summarized in Table~\ref{tab:related-work}.
Since event cameras are novel sensors (commercialized since 2008), there are only a handful of papers on SL systems.
These can be classified according to whether the shape of the light source (point, line, 2D pattern) 
and according to the number of event cameras used.

One of the earliest works combined a DVS with a pulsed laser line to reconstruct a small terrain~\cite{Brandli13fns}. 
The pulsed laser line was projected at a fixed angle with respect to the DVS while the terrain moved beneath, perpendicular to the projected line.
The method used an adaptive filter to distinguish the events caused by the laser (up to $f\!\sim$\SI{500}{\Hz}) from the events caused by noise or by the terrain's motion.

\begin{figure}
    \centering
    \includegraphics[trim={0 0 6.5cm 0.4cm},clip,width=.8\linewidth]{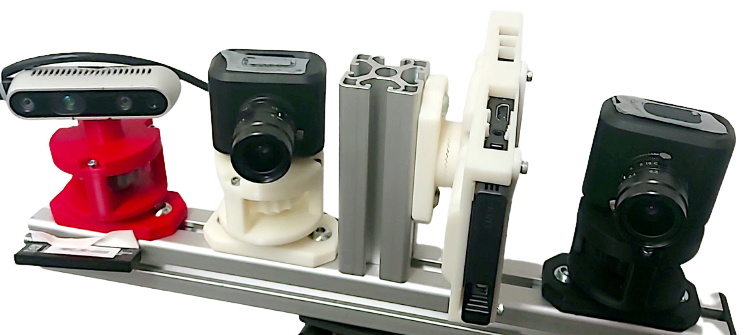}
    \vspace{-1ex}
    \caption{\emph{Physical setup used in the experiments} (Sect.~\ref{sec:experim:hardware}).
    From left to right: Intel RealSense, Prophesee Gen3S1.1 event camera and Sony Mobile point projector MPCL1A.
    Our method is designed for the point projector and the event camera, here separated by a \SI{11}{\centi\meter} baseline.
    The Intel RealSense D435 sensor is only used for comparison.
    }
    \label{fig:setup}
\end{figure}
\begin{table}[t]
    \centering
    \begin{adjustbox}{max width=\columnwidth}
    \setlength{\tabcolsep}{3pt}
    \begin{tabular}{llll}
    \toprule
    \textbf{Method}  &  \textbf{Event camera(s)} & \textbf{(pixels)} & \textbf{Projector} \\ \midrule
    Brandli \cite{Brandli13fns} & DVS128 & $128 \times 128$ & Laser line \SI{500}{\Hz}\\
    Matsuda \cite{Matsuda15iccp} & DVS128 & $128 \times 128$ & Laser point 60 fps\\
    Leroux \cite{Leroux18arxiv} & ATIS & $304 \times 240$ & DLP TI LightCrafter 3000\\
    Mangalore \cite{Mangalore20spl} & DAVIS346 & $346 \times 260$ & DLP TI LightCrafter 4500\\
    Martel \cite{Martel18iscas} & Stereo DAVIS240 & $240 \times 180$ & Laser beam \\
    \textbf{Ours} & Prophesee Gen3S1.1 & $640 \times 480$ & Laser point 60 fps\\
    \bottomrule
    \end{tabular}
    \end{adjustbox}
    \vspace{-1ex}
    \caption{\label{tab:related-work}Summary of event-camera--based structured-light depth estimation works.}
    \vspace{-2ex}
\end{table}

The SL system MC3D~\cite{Matsuda15iccp} comprised a laser point projector (operating up to \SI{60}{\Hz}) and a DVS. 
The laser raster-scanned the scene, and its reflection was captured by the DVS, which converted temporal information of events at each pixel into disparity.
It exploited the redundancy suppression and high temporal resolution of the DVS, also showing appealing results in dynamic scenes.
{In \cite{Leroux18arxiv}, a Digital Light Processing (DLP) projector was used to illuminate the scene with frequency-tagged light patterns.
Each pattern's unique frequency facilitated the establishment of correspondences between the patterns and the events,
leading to a sparse depth map that was later interpolated.

Recently, \cite{Martel18iscas} combined a laser light source with a \emph{stereo} setup consisting of two 
DAVIS event cameras~\cite{Brandli14ssc}.
The laser illuminated the scene and the synchronized event cameras recorded the events generated by the reflection from the scene.
Hence the light source was used to generate stereo point correspondences, which were then triangulated (back-projected) to obtain a 3D reconstruction.

More recently, \cite{Mangalore20spl} proposed a SL system with a fringe projector and an event camera. 
A sinusoidal 2D pattern with different frequencies illuminated the scene and its reflection was captured by the camera and processed (by phase unwrapping) to generated depth estimates.

The closest work to our method is MC3D~\cite{Matsuda15iccp} since both use a laser point-projector and a single event camera,
which is a sufficiently general and simple scenario that allows us to exploit the high-speed advantages of event cameras and the focusing power of a point light source.
In both methods we may interpret the laser and camera as a stereo pair.
The principle behind MC3D is to map the spatial disparity between the projector and event camera to temporal information of the events.
When events are generated, their timestamps are mapped to disparity by multiplying by the projector's scanning speed.
This operation amplifies the noise inherent in the event timestamps and leads to brittle stereo correspondences.
Moreover, this noise amplification depends on the projector's speed, which is product of the projector resolution and scanning frequency. Hence, MC3D's performance degrades as the scanning frequency increases.
By contrast, our method maximizes the spatio-temporal consistency between the projector's and event camera's data, thus leading to lower errors (especially with higher scanning frequencies).
By exploiting the regularities in neighborhoods of event data, as opposed to the point-wise operations in MC3D, our method improves robustness against noise.

\section{Depth Estimation}
\label{sec:methodolody}

This section introduces basic considerations of the event-camera - projector setup (Section~\ref{sec:method:preliminaries}) 
and then presents our optimization approach to depth estimation (Section~\ref{sec:method:energy-based-formulation}) 
using spatio-temporal consistency between the signals used on the event camera and the projector.
Overall, our method is summarized in Fig.~\ref{fig:geometricConfig} and Algorithm~\ref{alg:pseudo-code-patches}.

\subsection{Basic Considerations}
\label{sec:method:preliminaries}
\begin{figure}[t]
    \centering
    \includegraphics[trim={0cm 0cm 0cm 0.8cm},clip,width=0.82\linewidth]{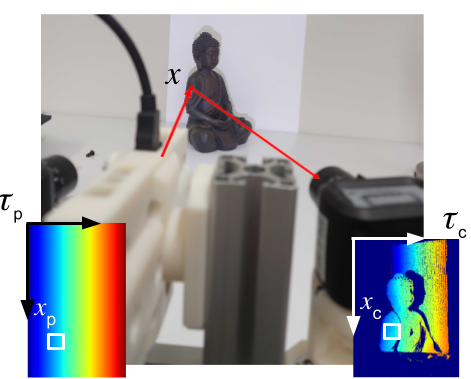}
    \vspace{-1ex}
    \caption{\emph{Geometry of the setup}.
    \mm{The laser projector illuminates the scene one point $\mathbf{x}$ at a time via a ray through point $\bxp$, which is recorded by the camera as an event at pixel $\bxc$.
    As justified in Section~\ref{sec:experim:hardware}, we adopt a vertical raster scan pattern (from top to bottom and left to right).
    The timestamps $\taup$ of a scan pass of the projector ($1/f$ seconds) and the timestamps of the corresponding events ($\tauc$) are displayed in pseudocolor, from blue (past) to red (recent).}
    The projector does not emit color patterns.
    }
    \label{fig:geometricConfig}
    \vspace{-1.5ex}
\end{figure}

We consider the problem of depth estimation using a laser point projector and an event camera.
Fig.~\ref{fig:geometricConfig} illustrates the geometry of our configuration.
The projector illuminates the scene by moving a laser light source in a raster scan fashion. %
The changes in illumination caused by the laser are observed by the event camera, whose pixels respond asynchronously by generating events\footnote{An event camera generates an event $e_k = (\bx_k,t_k,p_k)$ at time $t_k$ when the increment of logarithmic brightness at the pixel $\bx_k=(x_k,y_k)^\top$ 
reaches a predefined threshold $C$:
$L(\bx_k,t_k) - L(\bx_k,t_k-\Delta t_k) = p_k C$, where $p_k \in \{-1,+1\}$ is the sign (polarity) of the brightness change, 
and $\Delta t_k$ is the time since the last event at the same pixel location. \cite{Gallego20pami,Gallego15arxiv}}.
Ideally, every camera pixel receives light from a single scene point, which is illuminated by a single location of the laser as it sweeps through the projector's pixel grid (Fig.~\ref{fig:geometricConfig}).
Since the light source moves in a predefined manner and the event camera and the projector are synchronized, 
as soon as an event is triggered, one can match it to the current light source pixel (neglecting latency, light traveling time, etc.) 
to establish a stereo correspondence between the projector and the camera.
This concept of converting \emph{point-wise} temporal information into disparity was explored in~\cite{Matsuda15iccp}, 
which relied on precise timing of both laser and event camera to establish accurate correspondences.
However this one-to-one, ideal situation breaks down due to noise as the scanning speed increases.
Let us introduce the time constraints and effects from both devices: projector and event camera.

\textbf{Projector's Sweeping Time and Sensor's Temporal Resolution}.
Without loss of generality, we first assume the projector and the event camera are in a canonical stereo configuration, i.e., epipolar lines are horizontal (this can be achieved via calibration and rectification).
The time that it takes for the projector to move its light source from one pixel to the next one horizontally in the raster scan, $\delta t$, is inversely proportional to the scanning frequency $f$ and the projector's spatial resolution ($W \times H$ pixels):
\vspace{-0.5ex}
\begin{equation}
\label{eq:projector-time-dt}
\delta t = 1 / (f\,H\,W).
\vspace{-0.5ex}
\end{equation}
For example, our projector scans at $f=\SI{60}{\Hz}$ and has $1920 \times 1080$ pixels, 
thus it takes $1/f = \SI{16.6}{\milli\second}$ to sweep over all its pixels, spending at most $\delta t \approx \SI{8}{\nano\second}$ per pixel.
This is considerably smaller than the temporal resolution of event cameras (\SI{1}{\micro\second}) 
(i.e., the camera cannot perceive the time between consecutive raster-scan pixels).
To overcome this issue and be able to establish stereo correspondences along epipolar lines from time measurements, we take advantage of geometry: 
we \emph{rotate} the projector by \SI{90}{\degree}, so that the raster scan is now vertical (Fig.~\ref{fig:geometricConfig}).
The time that it now takes for the projector to move its light source from a pixel to its adjacent one on the still horizontal epipolar line is
\vspace{-0.5ex}
\begin{equation}
\label{eq:projector-time-dt-line}
\delta t_{\text{line}} = 1 / (f\,H),
\vspace{-0.5ex}
\end{equation}
the time that it takes to sweep over one of the $H$ raster lines.
For the above projector, $\delta t_{\text{line}}\approx \SI{15.4}{\micro\second}$, which is larger than the temporal resolution of most event cameras~\cite{Gallego20pami}.
Hence, the event camera is now able to distinguish between consecutive projector pixels on the same epipolar line.

\textbf{Event camera noise sources}: Let us describe the different noise characteristics of event cameras that factor into this system. 
\emph{Latency} is defined as the time it takes for an event to be triggered since the moment the logarithmic change of intensity exceeded the threshold.
Typically this latency can range from \SI{15}{\micro\second} to \SI{1}{\milli\second}.
Since this affects all the timestamps equally, this can be considered as a constant offset that does not affect relative timestamps between consecutive events.
\emph{Jitter} is the random noise that appears in the timestamps.
This can have a huge variance depending  on the scene and the illumination conditions. 
\emph{BurstAER mode}. This pixel read-out mode is very common in high resolution event cameras.
It is a technique used to quickly read events from the pixel array.
Instead of reading out each individual event pixel (which takes longer time for higher resolution cameras), this method reads out an entire row or group of rows together and assigns the same timestamp to all the events in these rows.
This causes banding effects that appear in the event timestamps,
hence they also affect the quality of the reconstructed depth map.

\subsection{Maximizing Spatio-Temporal Consistency}
\label{sec:method:energy-based-formulation}

The method in \cite{Matsuda15iccp}, (mentioned in Section~\ref{sec:method:preliminaries}, and described in Section~\ref{sec:experim:baseline} as baseline), computes disparity independently for each event and is, therefore, highly susceptible to noise, especially as the scanning speed increases.
We now propose a method that processes events in space-time neighborhoods 
and exploits the regularity of surfaces present in natural scenes 
to improve robustness against noise and produce spatially coherent 3D reconstructions.

\textbf{Time Maps}:
The laser projector illuminates the scene in a raster scan fashion. %
During one scanning interval, $T = 1/f$, the projector traverses each of its pixels $\bxp$ at a precise time $\taup$, 
which allows us to define a time map over the projector's pixel grid: $\bxp \mapsto \taup(\bxp)$.
Similarly for the event camera we can define another time map (see \cite{Lagorce17pami}) $\bxc \mapsto \tauc(\bxc)$, 
where $\tauc(\bxc)$ records the timestamp of the last event at pixel $\bxc$.
Owing to this similarity between time maps and the fact that the projector emits light whereas the camera acquires it, 
we think of the projector as an ``inverse'' event camera.
That is, the projector creates an ``illumination event'' $\tilde{e}=(\bxp,t_p,1)$ when light at time $t=t_p$ traverses pixel $\bxp$. 
These ``illumination events'' are sparse, follow a raster-like pattern and are $\delta t \approx \SI{8}{\nano\second}$ apart \eqref{eq:projector-time-dt}.

For simplicity, we do not make a distinction between $\taup$ and $\tauc$ and refer to them as time maps (i.e., regardless of whether they are in the projector's or the event camera's image plane).
Exemplary time maps are shown in \Fig \ref{fig:geometricConfig}.

\textbf{Geometric Configuration}:
A point $\bxc$ on the event camera’s image plane transfers onto a point $\bxp$ on the projector's image plane following a chain of transformations that involves the surface of the objects in the scene (Fig.~\ref{fig:geometricConfig}). 
If we represent the surface of the objects using the depth $Z$ with respect to the event camera, we have:
\vspace{-0.5ex}
\begin{equation}
\label{eq:transferPointFromCamera}
\bxp = \pi_p \bigl(\TE\; \pi_c^{-1} \bigl(\bxc, Z(\bxc)\bigr)\bigr)
\vspace{-0.5ex}
\end{equation} 
where $\pi_p$ is the perspective projection on the projector's frame, 
$\TE$ is the rigid-body motion from the camera to the projector,
$\pi_c^{-1}$ is the inverse perspective projection of the event camera 
(assumed to be well-defined by a unique point of intersection between the viewing ray from the camera and the surfaces in the scene).

\textbf{Time Constancy Assumption}. 
In the above geometric configuration, 
the ``illumination events'' from the projector induce regular events on the camera.
Equivalently, in terms of timestamps, 
the time map $\taup$ on the projector’s image plane induces a time map $\tauc$ on the camera’s image plane:
\vspace{-0.5ex}
\begin{equation}
\label{eq:IdealTimeSurfaceTransfer}
\tauc (\bxc) = \taup (\bxp).
\vspace{-0.5ex}
\end{equation} 
This equation states a \emph{time-consistency principle} between $\tauc, \taup$, which assumes negligible travel time and photoreceptor delay~\cite{Zhou18eccv,Ieng18fnins,Zhou20tro}, i.e., instantaneous transmission from projector to camera, as if ``illumination events'' and regular events were simultaneous. 
This time-consistency principle will play the same role that photometric consistency (e.g., the brightness constancy assumption $I_2(\bx_2) = I_1(\bx_1))$ plays in conventional (i.e., passive) multi-view stereo.

\textbf{Disparity map from stereo matching}. 
We formulate the problem of depth estimation using epipolar search, where we compare local neighborhoods of ``illumination'' and regular events (of size $W \!\times W \!\times T$) on the rectified image planes, seeking to maximize their consistency. 

In terms of time maps, a neighborhood $\tau_\star(\bx_\star, W)$, of size $W \times W$ pixels around point $\bx_\star$, is a compact representation of the spatio-temporal neighborhood of the point $\bx_\star$, since it not only contains spatial information but also temporal one, by definition of $\tau_\star$.
Our goal becomes then to maximize consistency~\eqref{eq:IdealTimeSurfaceTransfer}, and we do so by searching for $\taup(\bxp, W)$ (along the epipolar line) that minimizes the error
\begin{equation}
\label{eq:objectiveFunction}
Z^\ast \doteq \arg\min_{Z} C(\bxc, Z),
\end{equation}
\begin{equation}
\label{eq:residualCalculation}
C(\bxc, Z) \doteq \| \tauc(\bxc, W) - \taup(\bxp, W) \|^{2}_{L^2(W\times W)}.
\end{equation}

\begin{algorithm}[t]
\emph{Input}: Time maps $\taup,\tauc$ during one scanning interval,
and calibration (intrinsic and extrinsic parameters, $\TE$).\\
\emph{Output}: Depth map on the event camera image plane\\ 
\emph{Procedure}:\\ 
Initialize depth map $Z(\bxc)$ (using epipolar search along the epipolar line in the rectified projector plane)\\
  \For {each pixel $\bxc$}{
  Find $Z^{\ast}(\bxc)$ (i.e., the corresponding pixel $\bxp$ in~\eqref{eq:transferPointFromCamera}) 
  that minimizes~\eqref{eq:residualCalculation}.
 }
[Optional] Regularize the depth map using total variation (TV) denoising.
\caption{Depth estimation by spatio-temporal consistency maximization on local neighborhoods.}
\label{alg:pseudo-code-patches}
\end{algorithm}

\textbf{Discussion of the Approach}.
The temporal noise characteristics of event cameras (e.g jitter, latency, BurstAER mode, etc.) influence the quality of the obtained depth maps.
The advantages of the proposed method are as follows.
(\emph{i}) \emph{Robustness to noise} (event jitter): 
By considering spatio-temporal neighborhoods of events for stereo matching, our method becomes less susceptible to individual event's jitter than point-wise methods~\cite{Matsuda15iccp}.
(\emph{ii}) \emph{Less data required}: 
Point-wise methods improve depth accuracy on static scenes by averaging depth over multiple scans~\cite{Matsuda15iccp}. 
Our method exploits spatial relationships between events, which makes up for temporal averaging, and therefore produces good results with less data, thus enabling better reconstructions of dynamic scenes.
We may further smooth the depth maps by using a non-linear refinement step.
(\emph{iii}) \emph{Single step stereo triangulation}: Depth parametrization and stereo matching are combined in a single step, as opposed to the classical two-step approach of first establishing correspondences and then triangulating depth like SGM or SGBM. 
This improves accuracy by removing triangulation errors from non-intersecting rays.
(\emph{iv}) \emph{Trade-off controllability}: 
Parameter $W$ allows us to control the quality of the estimated depth maps, with a trade-off:
a small $W$ produces fine-detailed but noisy depth maps,
whereas a large $W$ filters out noise at the expense of recovering fewer details, with (over-)smooth depth maps.
Noise due to BurstAER mode or temporal resolution may affect large pixel areas.
We may mitigate this type of noise by using large neighborhoods at the expense of smoothing depth discontinuities.
On the downside, the method is computationally more expensive than~\cite{Matsuda15iccp}, albeit it is still practical.

The pseudo-code of the method is given in Alg.~\ref{alg:pseudo-code-patches}.
Overall, Alg.~\ref{alg:pseudo-code-patches} may be interpreted as a principled non-linear method to recover depth from raw measurements, which may be initialized by a simpler method, such as~\cite{Matsuda15iccp}.

\section{Experiments}
\label{sec:experim}
This section evaluates the performance of our event-based SL system for depth estimation.
We first introduce the hardware setup (Section~\ref{sec:experim:hardware}) 
and the baseline methods and ground truth used for comparison (Section~\ref{sec:experim:baseline}).
Then we perform experiments on static scenes to quantify the accuracy of Alg.~\ref{alg:pseudo-code-patches}, 
and on dynamic scenes to show its high-speed acquisition capabilities (Section~\ref{sec:experim:results}).

\subsection{Hardware Setup}
\label{sec:experim:hardware}
To the best of our knowledge, there is no available dataset on which the proposed method can be tested. 
Therefore, we build our setup using a Prophesee event camera and a laser point source projector (Fig.~\ref{fig:setup}).

\textbf{Event Camera}: 
In our setup, we use a Prophesee Gen3 camera~\cite{Posch11ssc,propheseeevk}, with a resolution of $640 \times 480$ pixels.
This sensor provides only regular events (change detection, not exposure measurement) which are used for depth estimation. 
We use a lens with a field of view (FOV) of \SI{60}{\degree}. 

\textbf{Projector Source}:
We use a Sony Mobile projector MP-CL1A.%
The projector has a scanning speed of \SI{60}{\Hz} and a resolution of $1920\times 1080$ pixels.
During one scan (an interval of \SI{16}{\milli\second}), the point light source moves in a raster scanning pattern. %
The light source consists of a Laser diode (Class 3R), of wavelength \SIrange{445}{639}{\nano\meter}.
The event camera and the laser projector are synchronized via an external jack cable.
The projector's FOV is \SI{20}{\degree}.
The projector and camera are \SI{11}{\centi\meter} apart and their optical axes form a \SI{26}{\degree} angle. 

\textbf{Calibration}: 
We calibrate the intrinsic parameters of the event camera using a standard calibration tool (Kalibr \cite{Furgale13iros}) on the images produced after converting events to images using E2VID \cite{Rebecq19pami} when viewing a checkerboard pattern from different angles.
We calibrate the extrinsic parameters of the camera-projector setup and the intrinsic parameters of the projector using a standard tool for SL systems~\cite{Moreno12impvt}.

\begin{table*}[t]
    \centering
    \begin{adjustbox}{max width=\linewidth}
    \setlength{\tabcolsep}{4pt}
    {\small
    \begin{tabular}{lrrrrrrrrrrrrrrrrrr}
        \toprule
        Scene    & \multicolumn{2}{c}{David}  & \multicolumn{2}{c}{Heart}   & \multicolumn{2}{c}{Book-Duck} & \multicolumn{2}{c}{Plant}  & \multicolumn{2}{c}{City of Lights}  & \multicolumn{2}{c}{Cycle}  & \multicolumn{2}{c}{Room}  & \multicolumn{2}{c}{Desk-chair}   & \multicolumn{2}{c}{Desk-books}\\

        Mean depth & \multicolumn{2}{c}{\SI{50}{\centi\meter}}  & \multicolumn{2}{c}{\SI{50}{\centi\meter}}   & \multicolumn{2}{c}{\SI{49}{\centi\meter}} & \multicolumn{2}{c}{\SI{70}{\centi\meter}}  & \multicolumn{2}{c}{\SI{90}{\centi\meter}}  & \multicolumn{2}{c}{\SI{90}{\centi\meter}} & \multicolumn{2}{c}{\SI{393.45}{\centi\meter}}  & \multicolumn{2}{c}{\SI{171.92}{\centi\meter}}   &         \multicolumn{2}{c}{\SI{151.3}{\centi\meter}} \\
        
        \cmidrule(l{1mm}r{1mm}){2-3} \cmidrule(l{1mm}r{1mm}){4-5} \cmidrule(l{1mm}r{1mm}){6-7} 
        \cmidrule(l{1mm}r{1mm}){8-9} \cmidrule(l{1mm}r{1mm}){10-11} \cmidrule(l{1mm}r{1mm}){12-13} \cmidrule(l{1mm}r{1mm}){14-15} \cmidrule(l{1mm}r{1mm}){16-17} \cmidrule(l{1mm}r{1mm}){18-19}
        Metrics   &  FR $\uparrow$ & RMSE $\downarrow$ &  FR $\uparrow$ & RMSE $\downarrow$ &  FR $\uparrow$ & RMSE $\downarrow$ &  FR $\uparrow$ & RMSE $\downarrow$ &  FR $\uparrow$ & RMSE $\downarrow$ & FR $\uparrow$ & RMSE $\downarrow$ & FR $\uparrow$ & RMSE $\downarrow$ & FR $\uparrow$ & RMSE $\downarrow$ & FR $\uparrow$ & RMSE $\downarrow$ \\
        \midrule
        MC3D     &  0.68  &  26.84 & 0.72 &  25.83 & 0.71 &  28.47 & 0.62  &  30.38 & 0.61 &  37.84 & 0.37  & 41.84 & 0.09 & 346.45 & 0.20 & 166.10 & 0.20 & 126.05\\
        MC3D proc.  &  0.84  &  14.81 & 0.85 &  14.99 & 0.84 &  20.33 & 0.71  &  24.73 & 0.75 &  26.93 & 0.47  & 25.28 & 0.14 & 195.51 & \textbf{0.54} & 96.56 & \textbf{0.40} & 93.78\\ 
        SGM      &  0.49  &  1.19  & 0.50 &  0.86 & 0.61  &  10.31 & 0.69  &  5.19 & 0.53  &  8.38 &  0.4   & 10.25 &  0.13 & 192.52 & 0.22 & 37.35 & 0.17 & 6.88 \\
        SGM proc.   &  0.81  &  1.08  & 0.84 &  \textbf{0.54} & 0.80  &  7.30  & 0.82  &  5.21 & 0.75  &  6.76 &  0.58  & 16.31 &  0.15 & 192.23 & 0.22 & 36.94 & 0.20 & 7.30 \\
        Ours     &  0.94  &  0.50  & 0.96 &  0.57 & 0.85  &  1.43  & 0.89  &  1.98 & 0.80  &  1.23 &  0.65  &  1.19   &  0.26 & 161.36 & 0.38 & 33.82 & 0.30 & 4.69  \\
        Ours proc.  &  \textbf{0.96}  &  \textbf{0.46}  & \textbf{0.98} &  0.55 & \textbf{0.88}  &  \textbf{1.40}  & \textbf{0.91}  &  \textbf{1.97} & \textbf{0.87}  &  \textbf{1.17} &  \textbf{0.66}  &  \textbf{1.15} &  \textbf{0.28} & \textbf{161.34} & 0.41 & \textbf{33.79} & 0.31 & \textbf{5.14}\\ \bottomrule
    \end{tabular}
    }
    \end{adjustbox}
    \vspace{-1ex}
    \caption{
    \emph{Static scenes}: RMSE (\SI{}{\centi\meter}) and fill rate (FR, depth map completion) with respect to ground truth.
    (See Fig.~\ref{fig:experim:static}).
    }
    \label{tab:comparison}
    \vspace{-1ex}
\end{table*}

\global\long\def\figWidth{0.141\linewidth}
\begin{figure}[t]
	\centering
    \setlength{\tabcolsep}{2pt}
	\begin{tabular}{
	M{0.35cm}
	M{\figWidth}
	M{\figWidth}
	M{\figWidth}
	M{\figWidth}
	M{\figWidth}}
		& Scene & GT \SI{1}{\second}  & MC3D \SI{16}{\milli\second} 
		& SGM \SI{16}{\milli\second} & \textbf{Ours \SI{16}{\milli\second}} 
		\\

		\rotatebox{90}{\makecell{Book-Duck}}
		&\includegraphics[width=\linewidth]{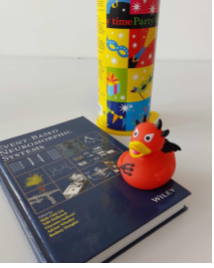}
		&\frame{\includegraphics[width=\linewidth]{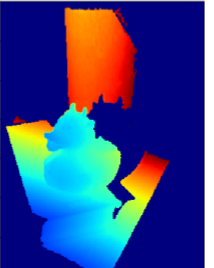}}
		&\frame{\includegraphics[width=\linewidth]{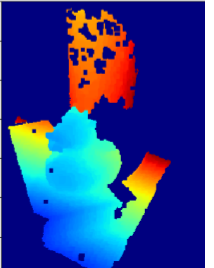}}
		&\frame{\includegraphics[width=\linewidth]{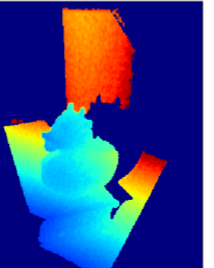}}
		&\frame{\includegraphics[width=\linewidth]{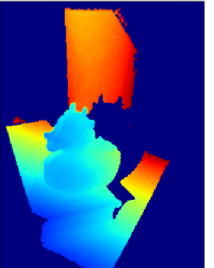}}
		\\

		\rotatebox{90}{\makecell{Plant}}
		&\includegraphics[trim={0 0 1cm 0},clip,width=\linewidth]{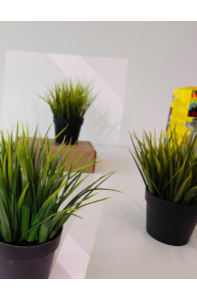}
		&\frame{\includegraphics[width=\linewidth]{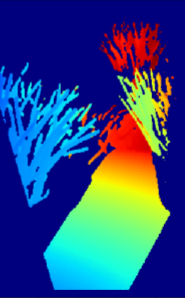}}
		&\frame{\includegraphics[width=\linewidth]{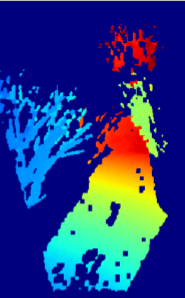}}
		&\frame{\includegraphics[width=\linewidth]{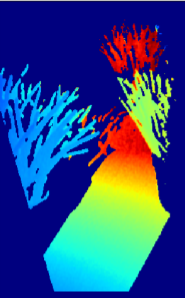}}
		&\frame{\includegraphics[width=\linewidth]{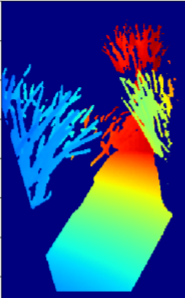}}
		\\

		\rotatebox{90}{\makecell{City Lights}}
		&\includegraphics[width=\linewidth]{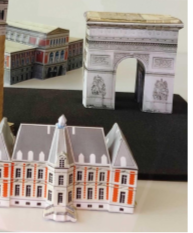}
		&\frame{\includegraphics[width=\linewidth]{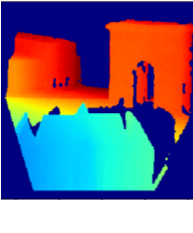}}
		&\frame{\includegraphics[width=\linewidth]{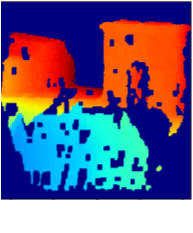}}
		&\frame{\includegraphics[width=\linewidth]{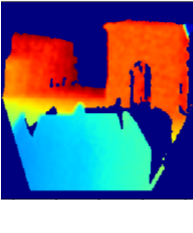}}
		&\frame{\includegraphics[width=\linewidth]{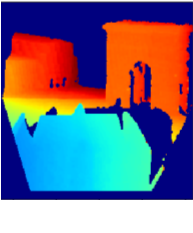}}
		\\

		\rotatebox{90}{\makecell{Cycle}}
		&\includegraphics[width=\linewidth]{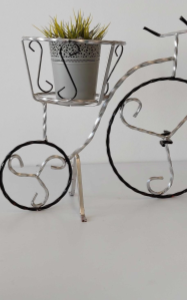}
		&\frame{\includegraphics[width=\linewidth]{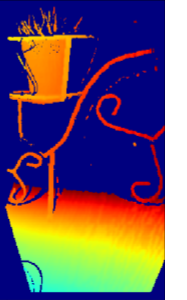}}
		&\frame{\includegraphics[width=\linewidth]{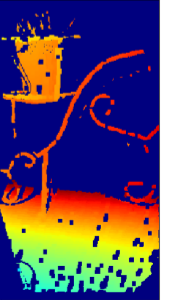}}
		&\frame{\includegraphics[width=\linewidth]{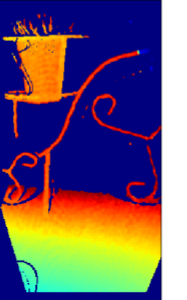}}
		&\frame{\includegraphics[width=\linewidth]{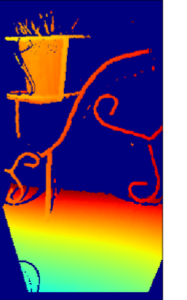}}
		\\

		\rotatebox{90}{\makecell{Room}}
		&\includegraphics[width=\linewidth]{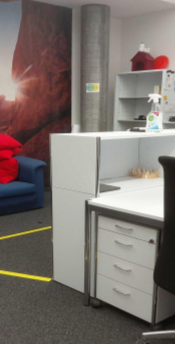}
		&\frame{\includegraphics[width=\linewidth]{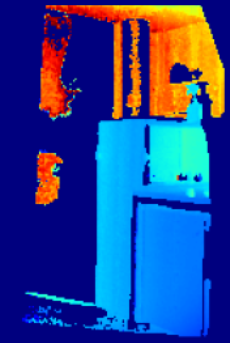}}
		&\frame{\includegraphics[width=\linewidth]{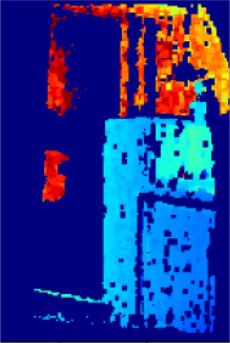}}
		&\frame{\includegraphics[width=\linewidth]{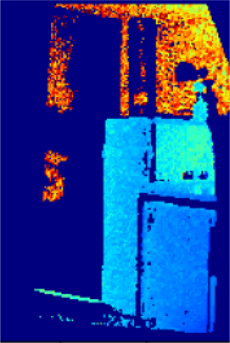}}
		&\frame{\includegraphics[width=\linewidth]{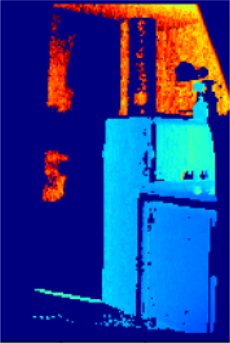}}
		\\

		\rotatebox{90}{\makecell{Desk-chair}}
		&\includegraphics[width=\linewidth]{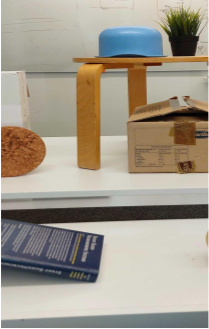}
		&\frame{\includegraphics[width=\linewidth]{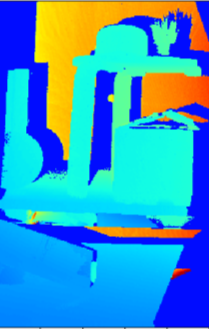}}
		&\frame{\includegraphics[width=\linewidth]{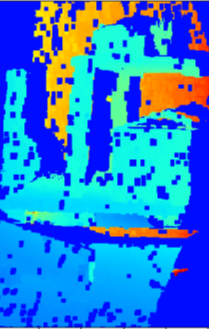}}
		&\frame{\includegraphics[width=\linewidth]{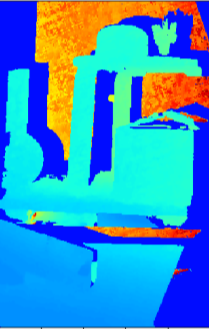}}
		&\frame{\includegraphics[width=\linewidth]{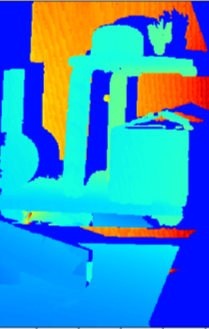}}
		\\
		
	    \rotatebox{90}{\makecell{Desk-books}}
		&\includegraphics[width=\linewidth]{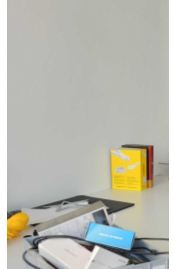}
		&\frame{\includegraphics[trim={0cm 0.5cm 0cm 0cm},clip,width=\linewidth]{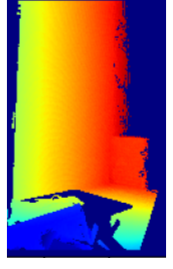}}
		&\frame{\includegraphics[trim={0cm 0.5cm 0cm 0cm},clip,width=\linewidth]{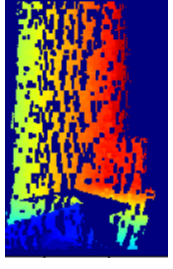}}
		&\frame{\includegraphics[trim={0cm 0.5cm 0cm 0cm},clip,width=\linewidth]{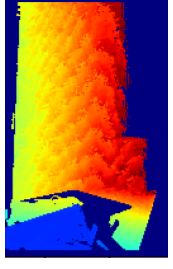}}
		&\frame{\includegraphics[trim={0cm 0.5cm 0cm 0cm},clip,width=\linewidth]{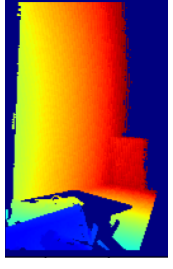}}
		\\
		
	\end{tabular}
    \vspace{-1ex}
	\caption{\emph{Static scenes}.
	Depth maps produced by several SL systems on static scenes.
	Our method (last column) produces, with 16ms acquisition time, almost as good results as ground truth (1s temporal averaging by MC3D).
	The zoomed-in insets in the even rows show how the different methods deal with smooth surfaces and fine details.
	}
	\label{fig:experim:static}
	\vspace{-2ex}
\end{figure}

\subsection{Baselines and Ground Truth}
\label{sec:experim:baseline}
Let us specify the depth estimation methods used for comparison and how ground truth depth is provided.

\textbf{MC3D Baseline}. 
We implemented the state-of-the-art method proposed in \cite{Matsuda15iccp}.
Moreover, we improved it by removing the need to scan the two end-planes of the scanning volume, which were used to linearly interpolate depth.
The details are described in the supplementary material.

Due to the event jitter of the event camera and noisy correspondences (e.g., missing matches), the disparity map for a single scanning period of \SI{16}{ms} is typically noisy and has many gaps (``holes'').
Hence, we apply a median filter in post-processing (also used by~\cite{Matsuda15iccp}).
However, this process does not remove all noise. %
Hence, we apply inpainting with hole filling and total variational (TV) denoising in post-processing.
In the experiments, we use as baseline the MC3D method~\cite{Matsuda15iccp} with a single single scan (\SI{16}{\milli\second}).

\textbf{SGM Baseline}. 
The main advantage of formulating the projector as an inverse event camera and its associated time map is that any stereo algorithm can be applied for disparity calculation between the projector' and event camera's time maps.
We therefore test the Semi-Global Matching (SGM) method~\cite{Hirschmuller08pami} on such timestamp maps.

\textbf{Ground truth}. We average the scans of MC3D over a period of \SI{1}{\second}.
With a frequency of \SI{60}{\Hz}, this temporal averaging approach combines $60$ depth scans into one.

\textbf{Evaluation metrics}.
We define two evaluation metrics: 
($i$) the root mean square error (\emph{RMSE}), namely the Euclidean distance between estimates and ground truth, measured in \SI{}{\centi\meter}, and ($ii$) the \emph{fill rate} (or completeness), namely the percentage of ground-truth points, which have been estimated by the proposed method within a certain error.
RMSE is often used to evaluate the quality of depth maps; 
however, this metric is heavily influenced by the scene depth, especially if there are missing points in the estimated depth map.
We therefore also measure the fill rate, with a depth error threshold of 1\% of the average scene depth.

\subsection{Results}
\label{sec:experim:results}
We assess the performance of our method on static and dynamic scenes, as well as in HDR illumination conditions.

\vspace{-1ex}
\subsubsection{Static Scenes}
\label{sec:experim:static-scenes}
Static scenes enable the acquisition of accurate ground truth by temporal averaging, which ultimately allows us to assess the accuracy of our method.
To this end, we evaluate our method on ten static scenes with increasing complexity:
a 3D printed model of Michelangelo's David, a 3D printed model of a heart, book-duck-cylinder, plants, City of Lights and cycle-plant.
We also include long-range indoor scenes of desk and room having maximum depth of \textbf{\SI{6.5} {\meter}}.
The scenes have varying depths (range and average depth).

Depth estimation results are collected in Fig.~\ref{fig:experim:static} and Table~\ref{tab:comparison}.
The depth error was measured on the overlapping region with the ground truth.
As it can be observed, on all scenes, our method, which processes the event data triggered by a single scan pass of the \SI{60}{\Hz} projector, outperforms the MC3D baseline method with the same input data (\SI{16}{\milli\second}).
Although SGM gives satisfactory results in comparison to MC3D, it suffers from artefacts that arise when temporal consistency is not strictly adhered to.
Table~\ref{tab:comparison} reports the fill rate (completion) and RMS error for our method and the two baselines (MC3D, SGM). 
The even rows incorporate post-processing (``proc''), which fills in holes (i.e., increases the fill ratio) and decreases the RMS depth error. The best results are obtained using our method and post-processing. 
However, the effect of post-processing is marginal in our method compared to the effect it has on the baseline methods.
\begin{figure}
    \centering
    {\includegraphics[trim={5.9cm 6cm 4.5cm 6.4cm},clip,width=0.79\linewidth]{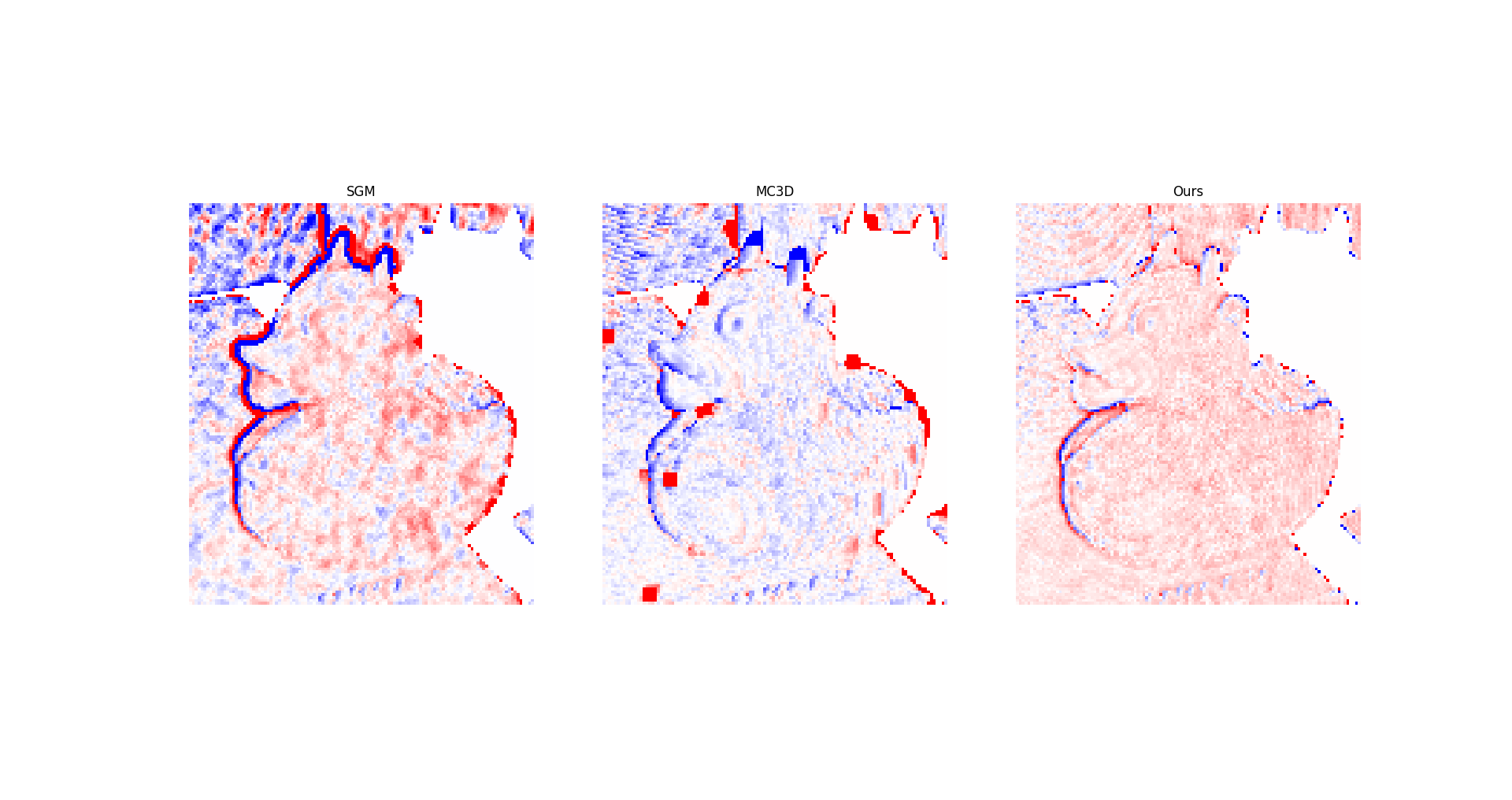}}
    \caption{\label{fig:depthdiff:signed}
    \up{Signed difference depth (with respect to GT) on Book-Duck scene.
    Left to right: SGM, MC3D, and Ours.}}
    \vspace{-1ex}
\end{figure}

Fig.~\ref{fig:depthdiff:signed} zooms into the signed depth errors for the Book-Duck scene (top row in Fig.~\ref{fig:experim:static}). 
Here, SGM gives the largest errors, specially at the duck's edges; 
MC3D yields smaller errors, but still has marked object contours and gaps; 
finally, our approach has the smallest error contours.

\subsubsection{High Dynamic Range Experiments}
We also assess the performance of our method on a static scene under different illumination conditions (Fig.~\ref{fig:experim:static:hdr}), 
which demonstrates the advantages of using an event-based SL depth system over conventional-camera--based depth sensor like Intel RealSense D435.

\global\long\def\figWidthRef{0.2322\linewidth}
\global\long\def\figWidth{0.2160\linewidth}
\global\long\def\figWidthRealSense{0.2808\linewidth}

\begin{figure}[h!]
	\centering
    \setlength{\tabcolsep}{2pt}
	\begin{tabular}{
	M{0.35cm}
	M{\figWidthRef}
	M{\figWidth}
	M{\figWidthRealSense}}
		& Scene & \textbf{Ours} \SI{16}{\milli\second} & Intel RealSense
		\\
		\rotatebox{90}{\makecell{Ambient light}}
		&\includegraphics[width=\linewidth]{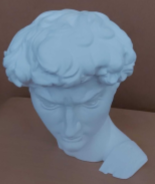}
		&\frame{\includegraphics[width=\linewidth]{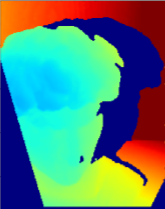}}
		&\frame{\includegraphics[width=\linewidth]{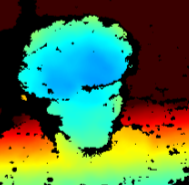}}
		\\
		
		\rotatebox{90}{\makecell{Lamp light}}
		&\includegraphics[width=\linewidth]{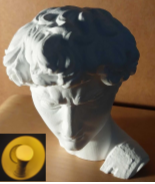}
		&\frame{\includegraphics[width=\linewidth]{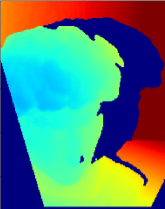}}
		&\frame{\includegraphics[width=\linewidth]{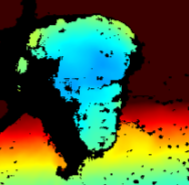}}
		\\
		
		\rotatebox{90}{\makecell{Floodlight}}
		&\includegraphics[width=\linewidth]{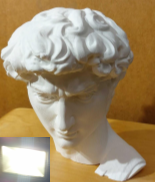}
		&\frame{\includegraphics[width=\linewidth]{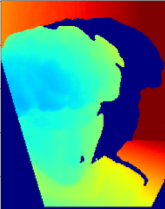}}
		&\frame{\includegraphics[width=\linewidth]{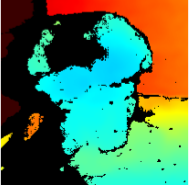}}
		\\
	\end{tabular}
	\vspace{-1ex}
	\caption{\emph{Effect of varying illumination conditions} (light intensity, HDR) on depth estimation. 
	See also Table~\ref{tab:comparison:hdr}.
	\label{fig:experim:static:hdr}}
\end{figure}
\begin{table}[h!]
    \centering
    \begin{adjustbox}{max width=.9\linewidth}
    \setlength{\tabcolsep}{3pt}
    {\small
    \begin{tabular}{lrrrrrr}
        \toprule
        Scene    & \multicolumn{2}{c}{Ambient light}  & \multicolumn{2}{c}{Lamp light}   & \multicolumn{2}{c}{Floodlight}  \\
        \cmidrule(l{1mm}r{1mm}){2-3} \cmidrule(l{1mm}r{1mm}){4-5} \cmidrule(l{1mm}r{1mm}){6-7} 
                  &  FR $\uparrow$ & RMSE $\downarrow$ &  FR $\uparrow$ & RMSE $\downarrow$ &  FR $\uparrow$ & RMSE $\downarrow$  \\
        \midrule
        MC3D       &  0.70 & 23.75 & 0.72 & 23.33 & 0.71 & 23.14 \\
        MC3D proc. &  0.90 & 10.67 & 0.92 & 9.84 & 0.92 & 10.26\\ 
        SGM        &  0.66 & 1.95 & 0.64 & 1.89 & 0.64 & 1.89 \\
        SGM proc.  &  0.90 & \textbf{1.89} & 0.86 & \textbf{1.83} & 0.86 & \textbf{1.83} \\ 
        Ours       &  \textbf{0.98} & 1.99 & \textbf{0.98} & 1.99 & \textbf{0.98} & 1.99 \\
        Ours proc. &  \textbf{0.98} & 1.98 & \textbf{0.98} & 1.98 & \textbf{0.98} & 1.98\\ \bottomrule
    \end{tabular}
    }
    \end{adjustbox}
    \vspace{-1ex}
    \caption{\label{tab:comparison:hdr}\emph{Effect of illumination conditions} on RMS error (\SI{}{\centi\meter}) and fill rate (FR, depth map completion).
    (See Fig.~\ref{fig:experim:static:hdr}).}
    \vspace{-3ex}
\end{table}

Fig.~\ref{fig:experim:static:hdr} shows qualitatively how our method provides consistent depth maps under different illumination conditions, whereas a frame-based depth sensor, e.g. Intel RealSense, does not cope well with such challenging scenarios.
Table~\ref{tab:comparison:hdr} compares our method against the event-based baselines in HDR conditions.
While all event-based methods estimate consistent depth maps across the HDR conditions, our method outperforms the MC3D baseline significantly.
We observe that as illumination increases, there is a slight decrease of the errors.
The reason is that the noise (i.e., jitter) in the event timestamps decreases with illumination.

\subsubsection{Sensitivity with respect to the Neighborhood Size}
\label{sec:experim:sensitivity}
Fig.~\ref{fig:experim:patch-sensitivity} qualitatively shows the performance of Alg.~\ref{alg:pseudo-code-patches} as the size of the local aggregation neighborhood increases from $W=3$ to $W=15$ pixels on the event camera's image plane.
As anticipated in Section~\ref{sec:method:energy-based-formulation}, there is a trade-off between accuracy, detail preservation, and noise reduction.
Our method allows us to control the desired depth estimation quality along this trade-off via the parameter $W$.
\global\long\def\figWidth{0.16\linewidth}
\global\long\def\figWidthBlue{0.25\linewidth}
\global\long\def\coorbarWidth{0.1\linewidth}
\begin{figure}[h!]
	\centering
    \setlength{\tabcolsep}{2pt}
	\begin{tabular}{
	M{0.1cm}
	M{\figWidth}
	M{\figWidthBlue}
	M{\figWidthBlue}
	M{\figWidthBlue}
	}
		& Object & $W=3$ & $W=7$ & $W=15$
		\\
		\rotatebox{90}{\makecell{ }}
		&\includegraphics[trim={0 0 0 1cm},clip,width=\linewidth]{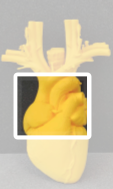}
		&\frame{\includegraphics[width=\linewidth]{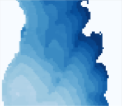}}
		&\frame{\includegraphics[width=\linewidth]{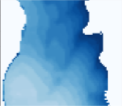}}
		&\frame{\includegraphics[width=\linewidth]{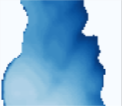}}
		\\[-2ex]%

	\end{tabular}
	\caption{\label{fig:experim:patch-sensitivity}
	\emph{Effect of the neighborhood size}: 
	Sensitivity with respect to the size of the local aggregation neighborhood, $W^2$ pixels.
	Depth maps obtained with Alg.~\ref{alg:pseudo-code-patches} using only one scan (\SI{16}{\milli\second}), for different values of the parameter $W$.
	There is a smoothness vs.~accuracy (detail preservation) trade-off that can be controlled by means of parameter $W$.
	}
	\vspace{-3ex}
\end{figure}

\subsubsection{Dynamic Scenes}
\label{sec:experim:motion-scenes}
\global\long\def\figWidth{0.172\linewidth}
\begin{figure}[t]
	\centering
    \setlength{\tabcolsep}{2pt}
	\begin{tabular}{
	M{0.35cm}
	M{\figWidth}
	M{\figWidth}
	M{\figWidth}
	M{\figWidth}
	M{\figWidth}}
		& Scene & MC3D \SI{16}{\milli\second} 
		& SGM \SI{16}{\milli\second} & \textbf{Ours} \SI{16}{\milli\second} & Intel Realsense
		\\

		\rotatebox{90}{\makecell{Tape spin}}
		&\includegraphics[width=\linewidth]{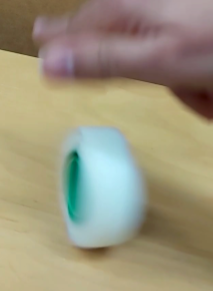}
		&\frame{\includegraphics[width=\linewidth]{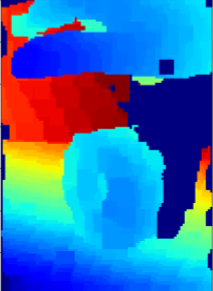}}
		&\frame{\includegraphics[width=\linewidth]{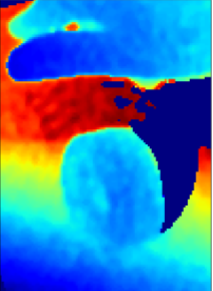}}
		&\frame{\includegraphics[width=\linewidth]{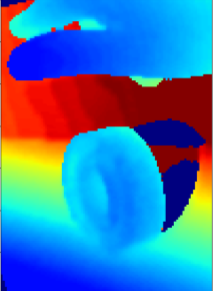}}
		&\frame{\includegraphics[trim={0 0 0 1.1cm},clip,width=\linewidth]{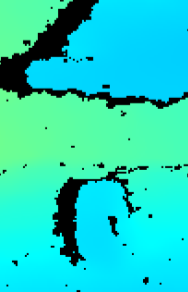}}
		\\
		
		\rotatebox{90}{\makecell{Fan}}
		&\includegraphics[width=\linewidth]{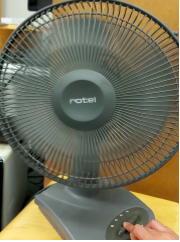}
		&\frame{\includegraphics[width=\linewidth]{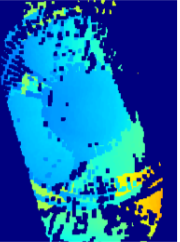}}
		&\frame{\includegraphics[width=\linewidth]{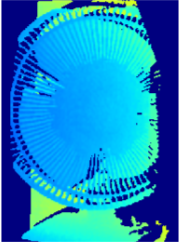}}
		&\frame{\includegraphics[width=\linewidth]{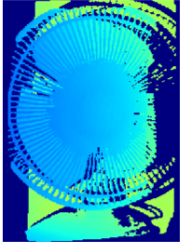}}
		&\frame{\includegraphics[width=\linewidth]{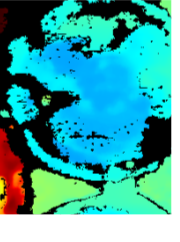}}
		\\

		\rotatebox{90}{\makecell{Multi-object}}
		&\includegraphics[width=\linewidth]{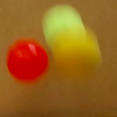}  %
		&\frame{\includegraphics[width=\linewidth]{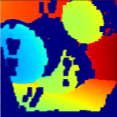}}
		&\frame{\includegraphics[width=\linewidth]{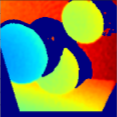}}
		&\frame{\includegraphics[width=\linewidth]{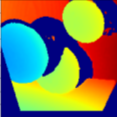}}
		&\frame{\includegraphics[width=\linewidth]{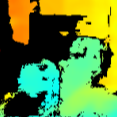}}
		\\
		\rotatebox{90}{\makecell{Table tennis}}
		&\includegraphics[width=\linewidth]{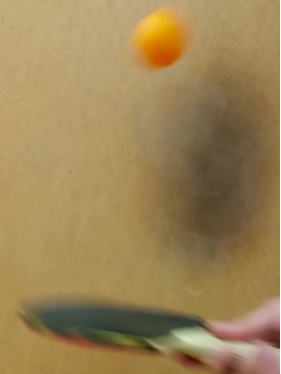}
		&\frame{\includegraphics[width=\linewidth]{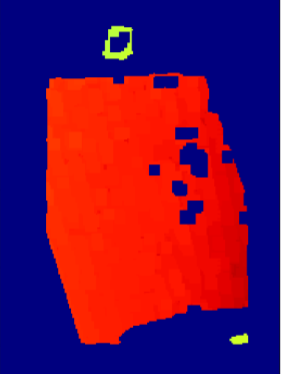}}
		&\frame{\includegraphics[width=\linewidth]{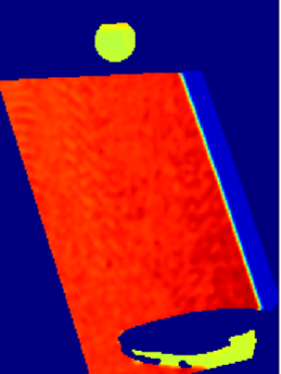}}
		&\frame{\includegraphics[width=\linewidth]{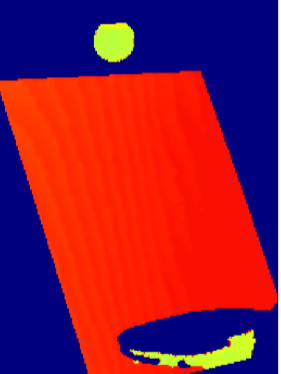}}
		&\frame{\includegraphics[width=\linewidth]{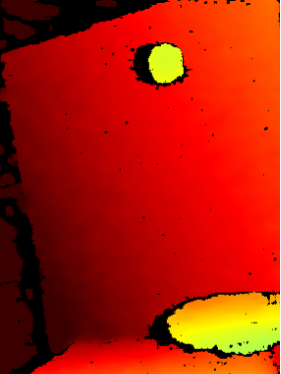}}
		\\
		
		\rotatebox{90}{\makecell{Origami fan}}
		&\includegraphics[width=\linewidth]{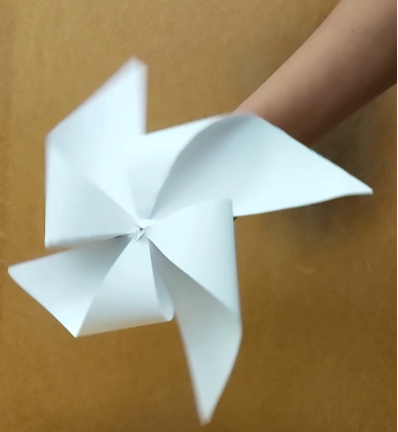}
		&\frame{\includegraphics[width=\linewidth]{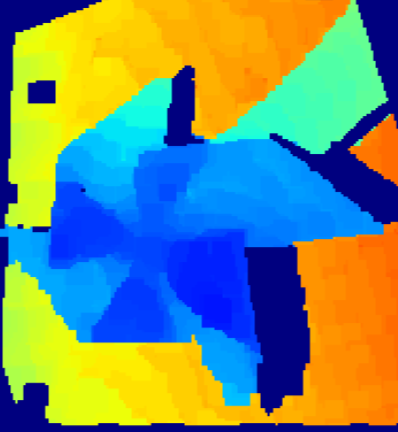}}
		&\frame{\includegraphics[width=\linewidth]{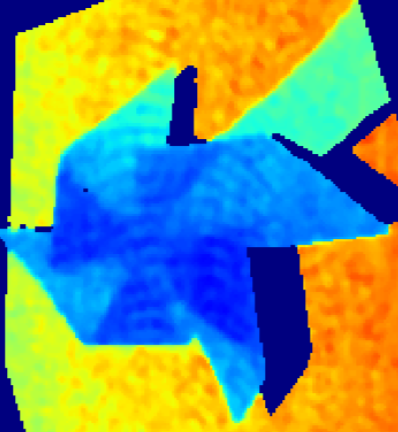}}
		&\frame{\includegraphics[width=\linewidth]{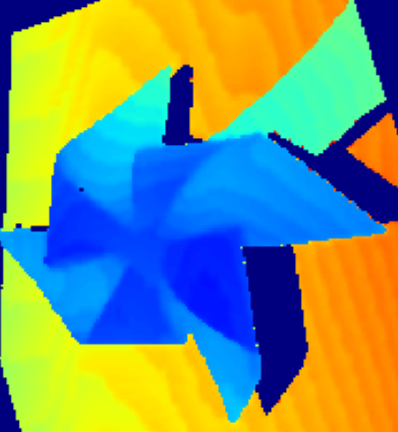}}
		&\frame{\includegraphics[width=\linewidth]{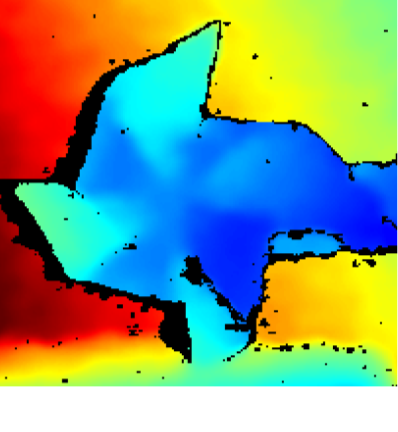}}
		\\
		
		\rotatebox{90}{\makecell{Air balloon}}
		&\includegraphics[trim={21cm, 0, 21cm, 0},clip,width=\linewidth]{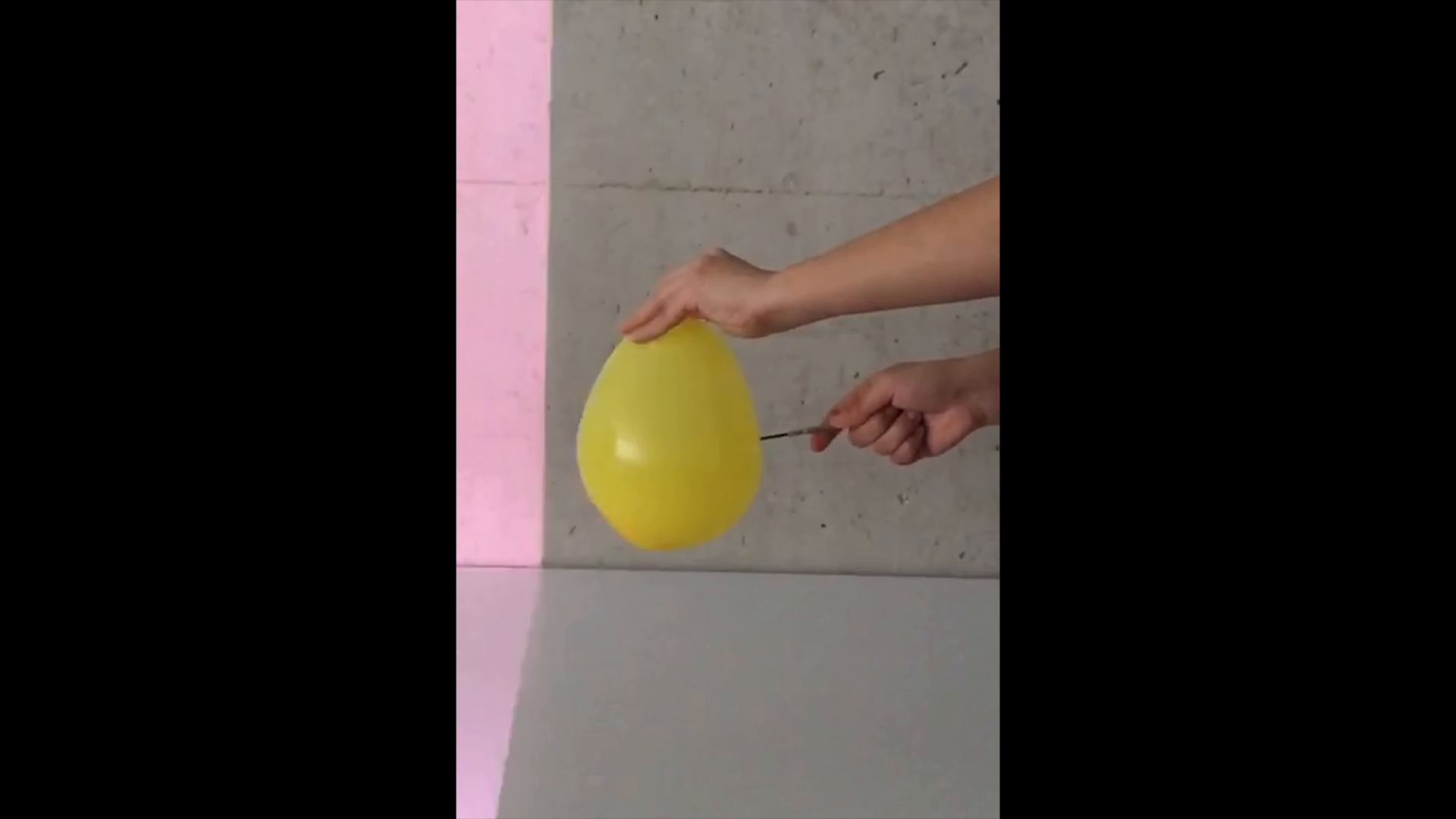}
		&\frame{\includegraphics[trim={5cm, 1.75cm, 5cm, 1.5cm},clip,width=\linewidth]{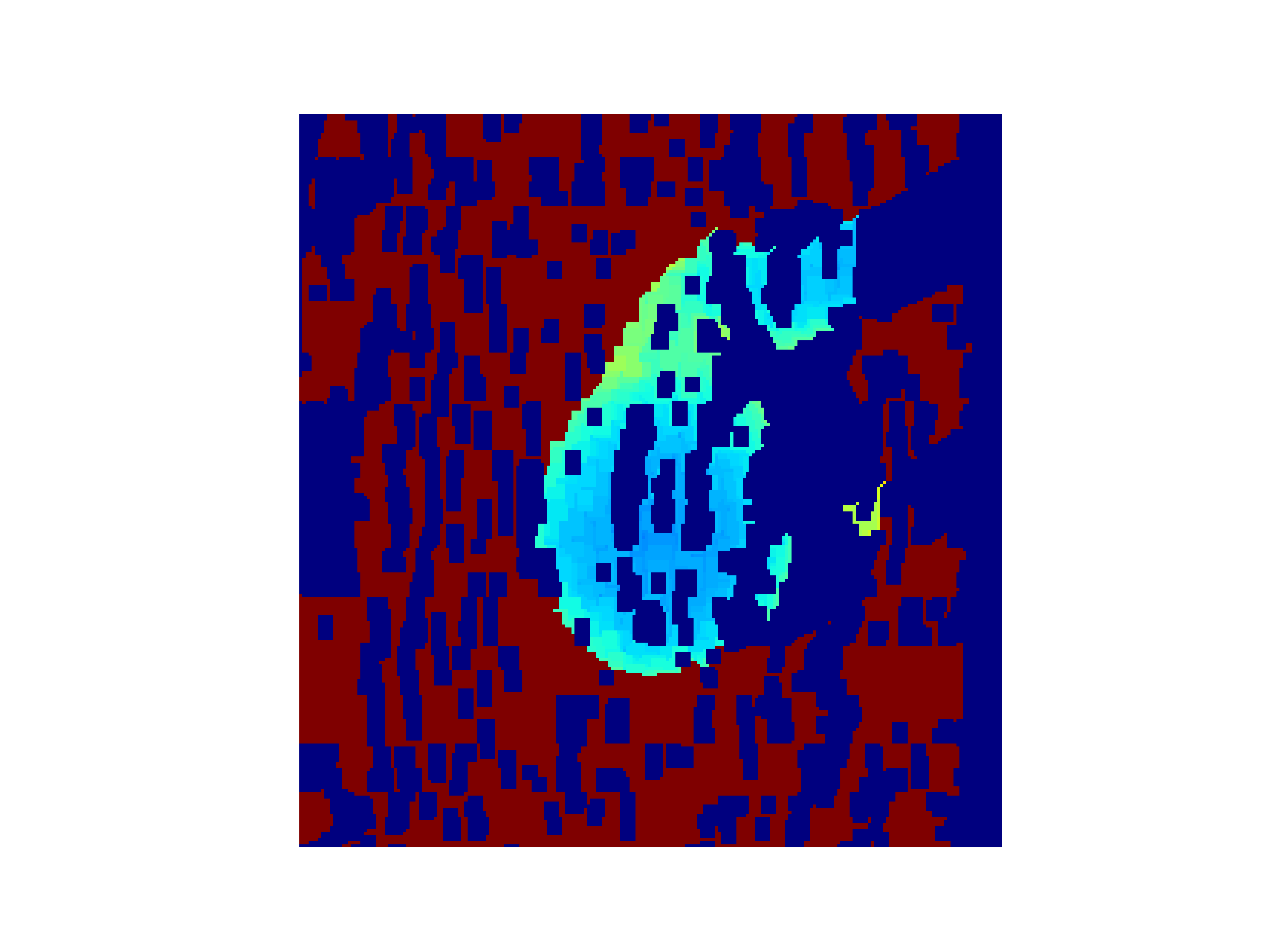}}
		&\frame{\includegraphics[trim={20cm, 2.7cm, 16.5cm, 5cm},clip,width=\linewidth]{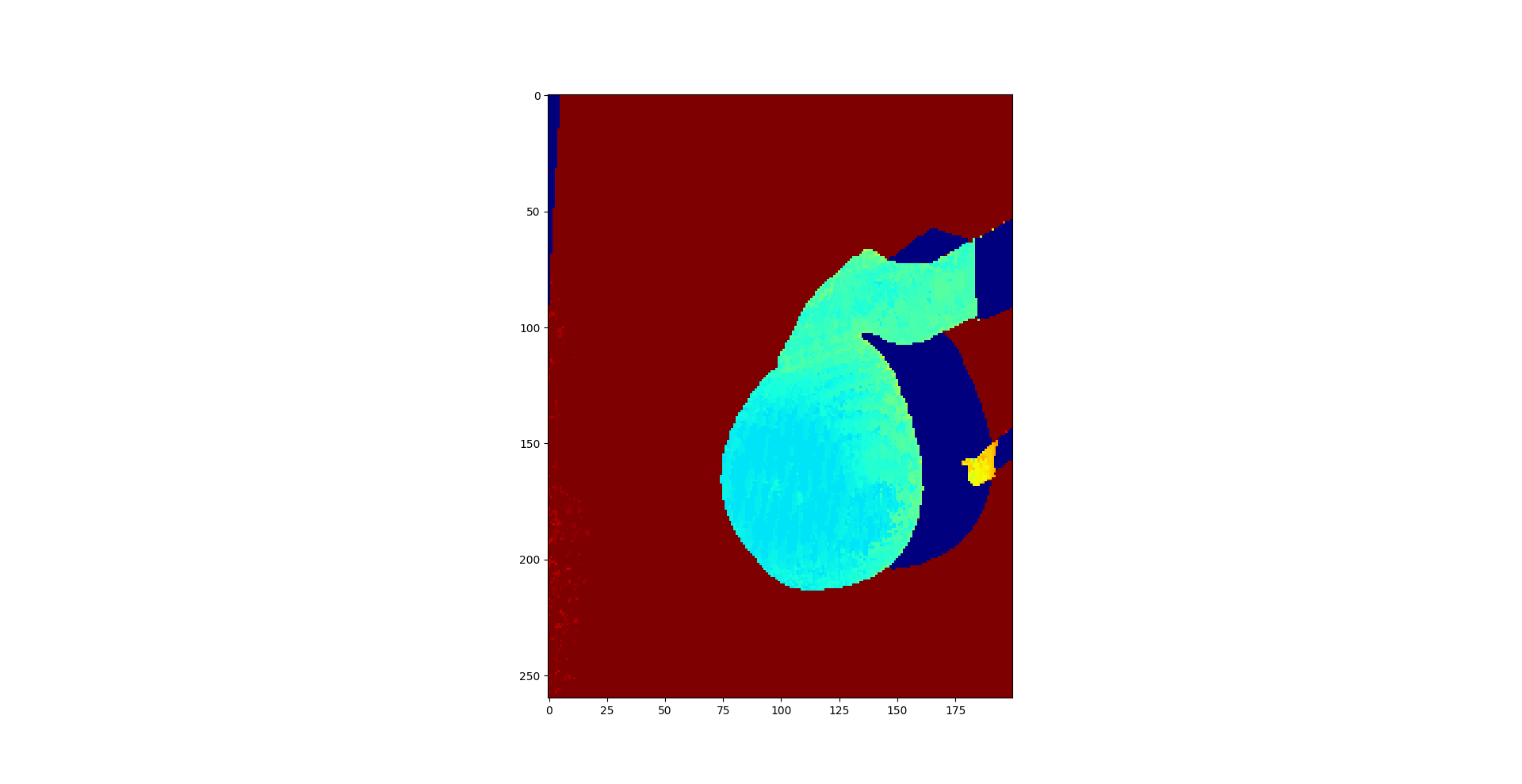}}
		&\frame{\includegraphics[trim={5cm, 1.75cm, 5cm, 1.5cm},clip,width=\linewidth]{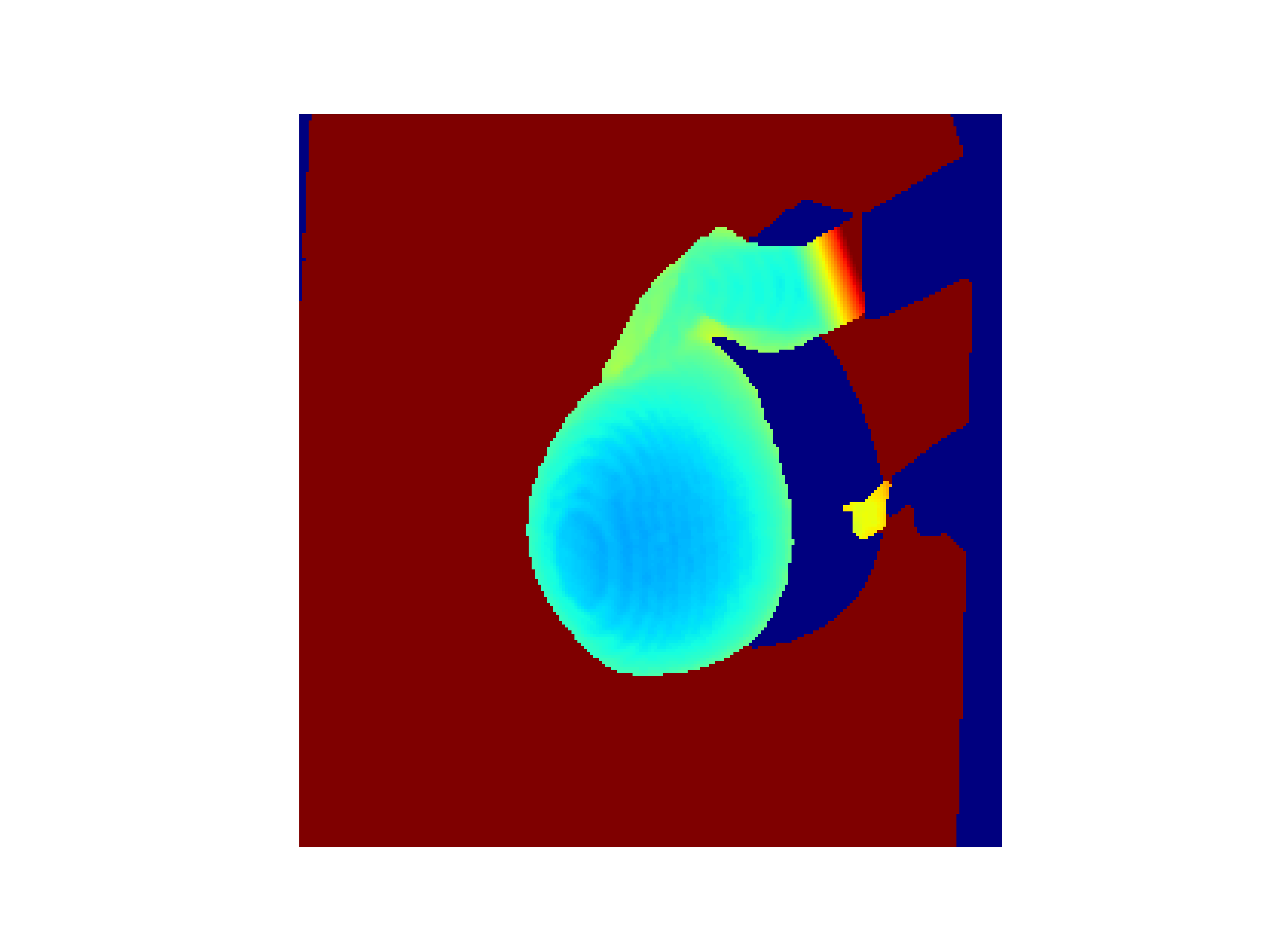}}
		&\frame{\includegraphics[trim={12cm, 3.75cm, 10cm, 3.5cm},clip,width=\linewidth]{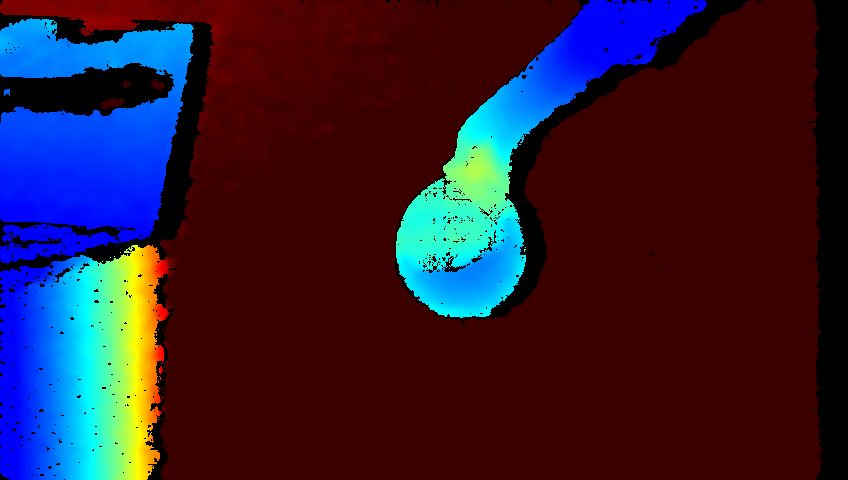}}
		\\
		
		\rotatebox{90}{\makecell{Water balloon}}
		&\includegraphics[trim={21cm, 4cm, 21cm, 0},clip,width=\linewidth]{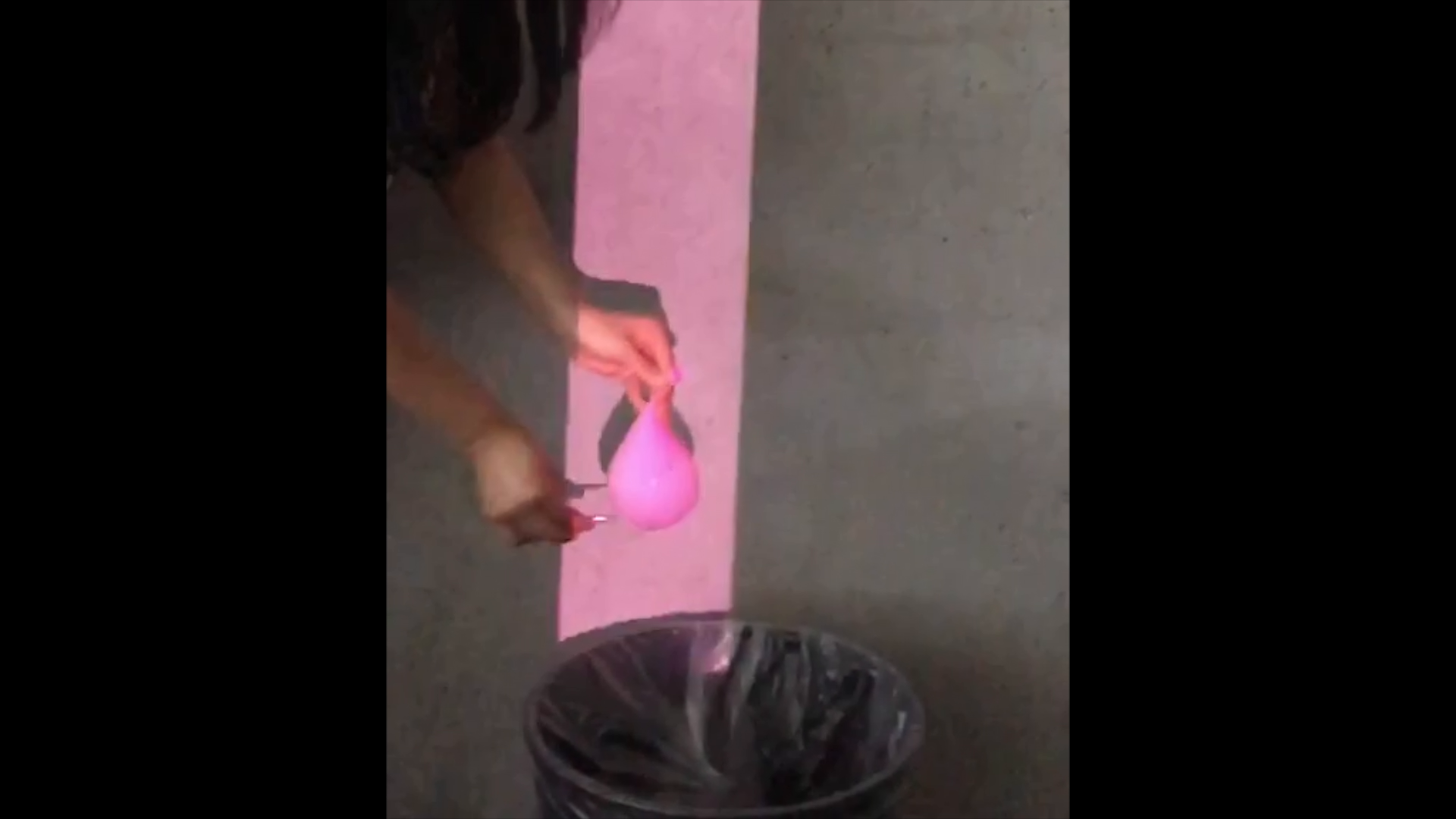}
		&\frame{\includegraphics[trim={6cm, 2.5cm, 5cm, 2.5cm},clip,width=\linewidth]{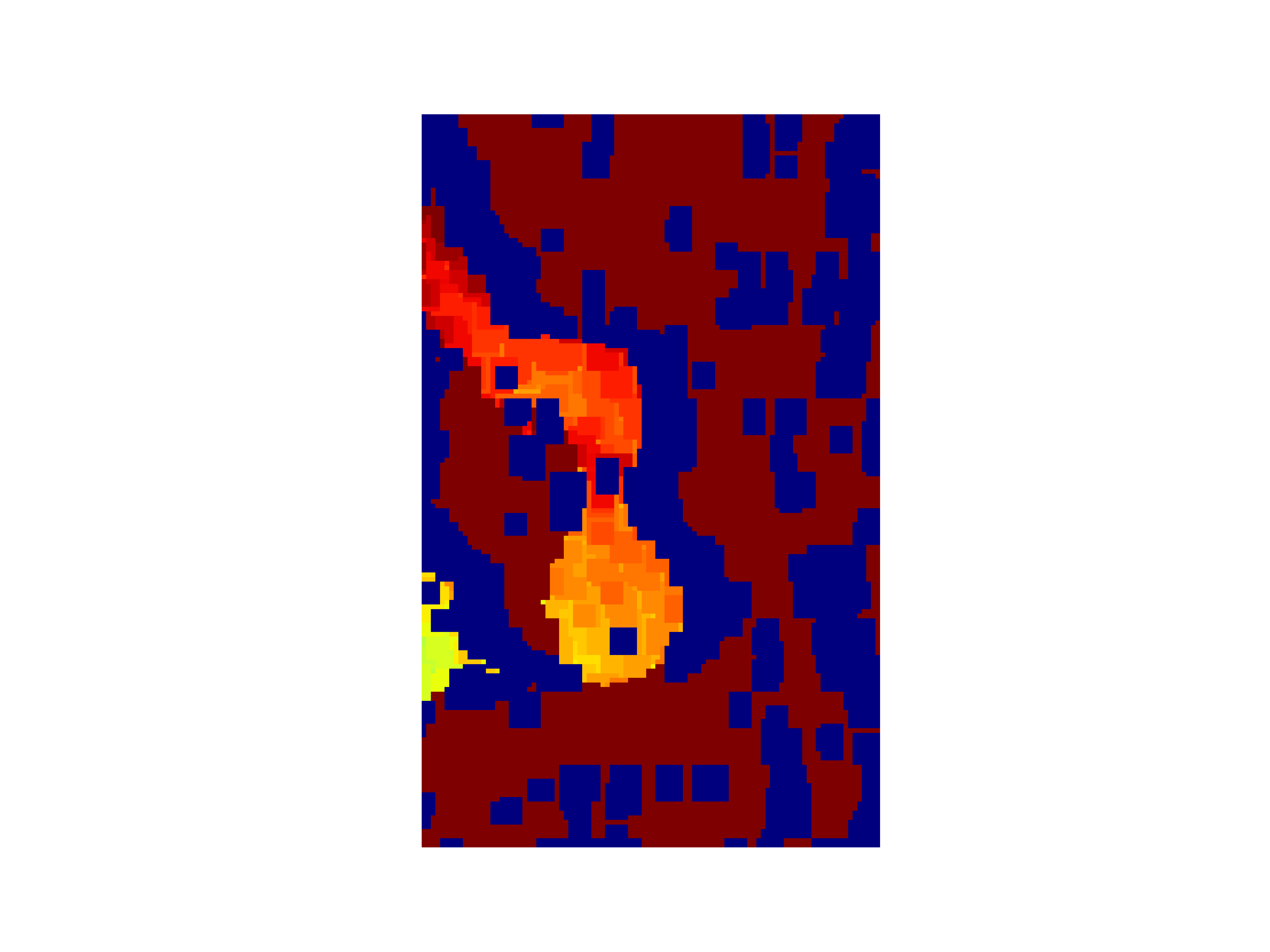}}
		&\frame{\includegraphics[trim={6.5cm, 2.5cm, 6cm, 4.5cm},clip,width=\linewidth]{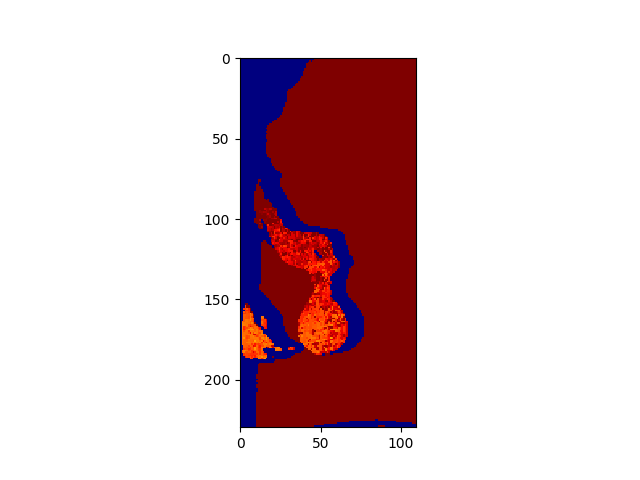}}
		&\frame{\includegraphics[trim={6cm, 2.5cm, 5cm, 2.5cm},clip,width=\linewidth]{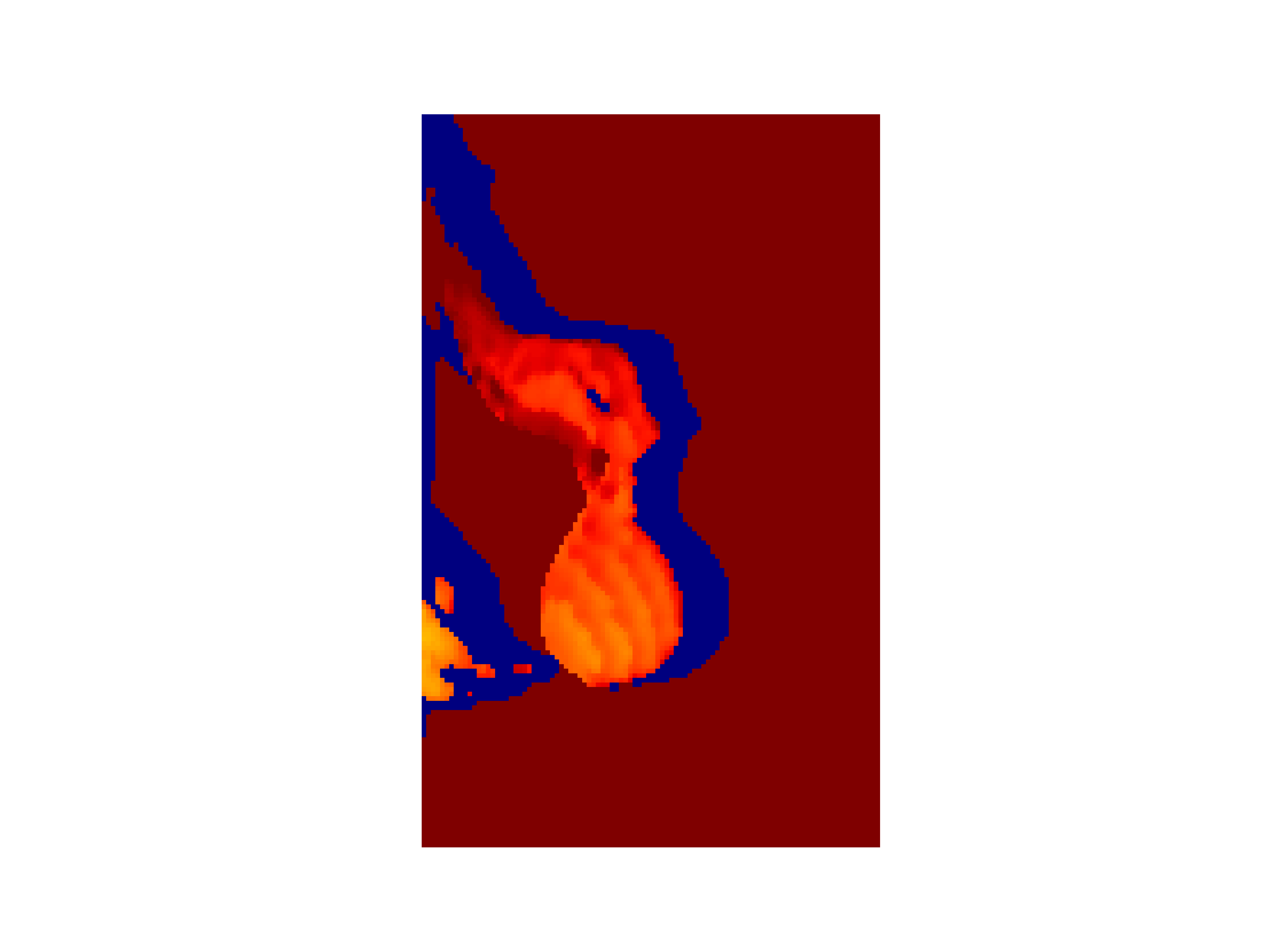}}
		&\frame{\includegraphics[trim={13cm, 2.5cm, 13.5cm, 10cm},clip,width=\linewidth]{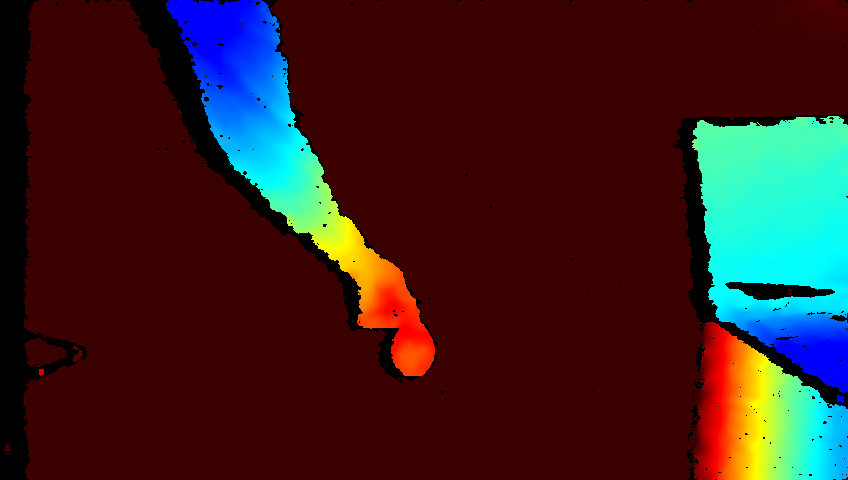}}
		\\
		
		\rotatebox{90}{\makecell{Coin}}
		&\includegraphics[trim={21cm, 0, 21cm,
		5cm},clip,width=\linewidth]{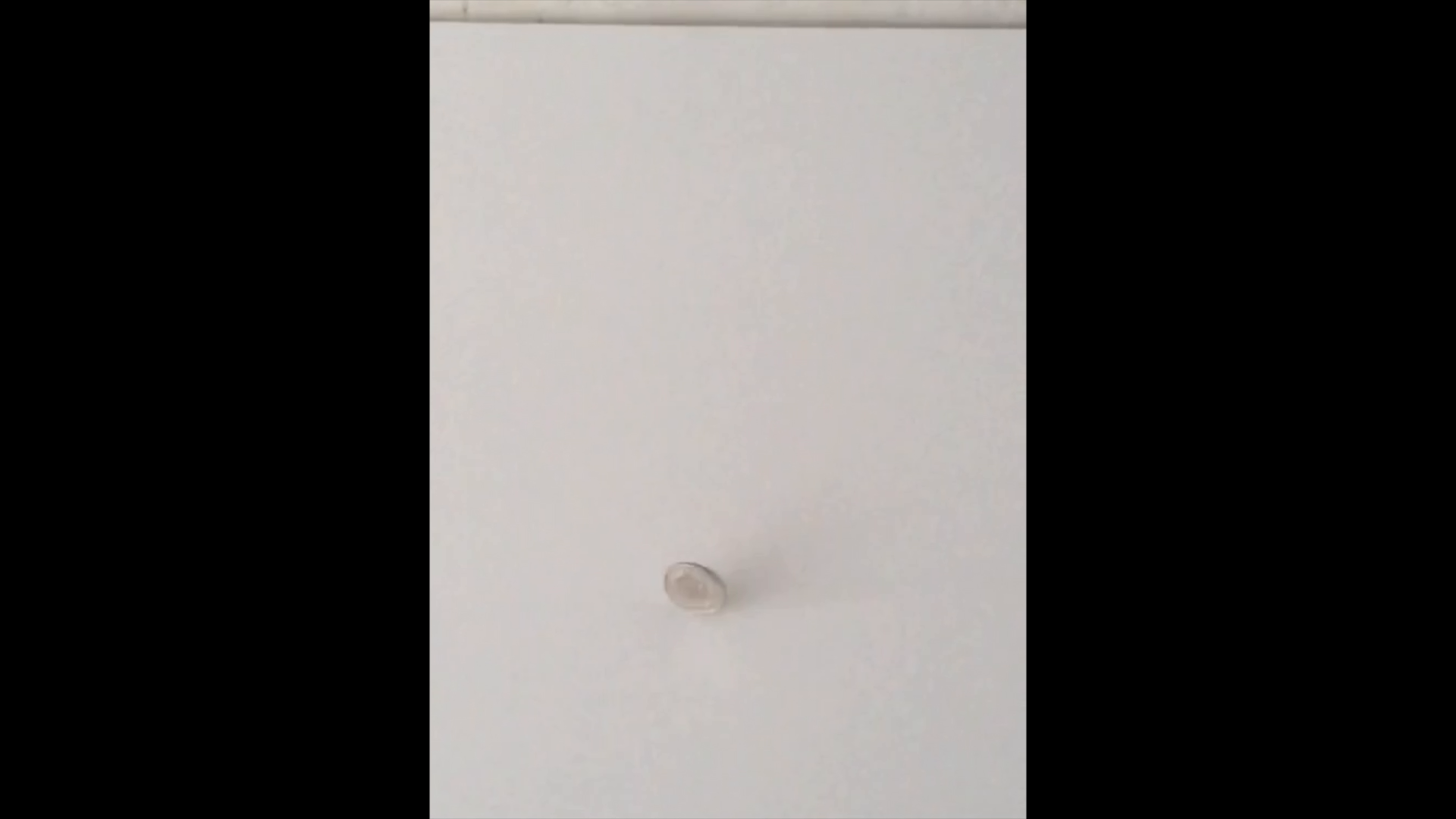}
		&\frame{\includegraphics[trim={6cm, 2.5cm, 5cm, 2.9cm},clip,width=\linewidth]{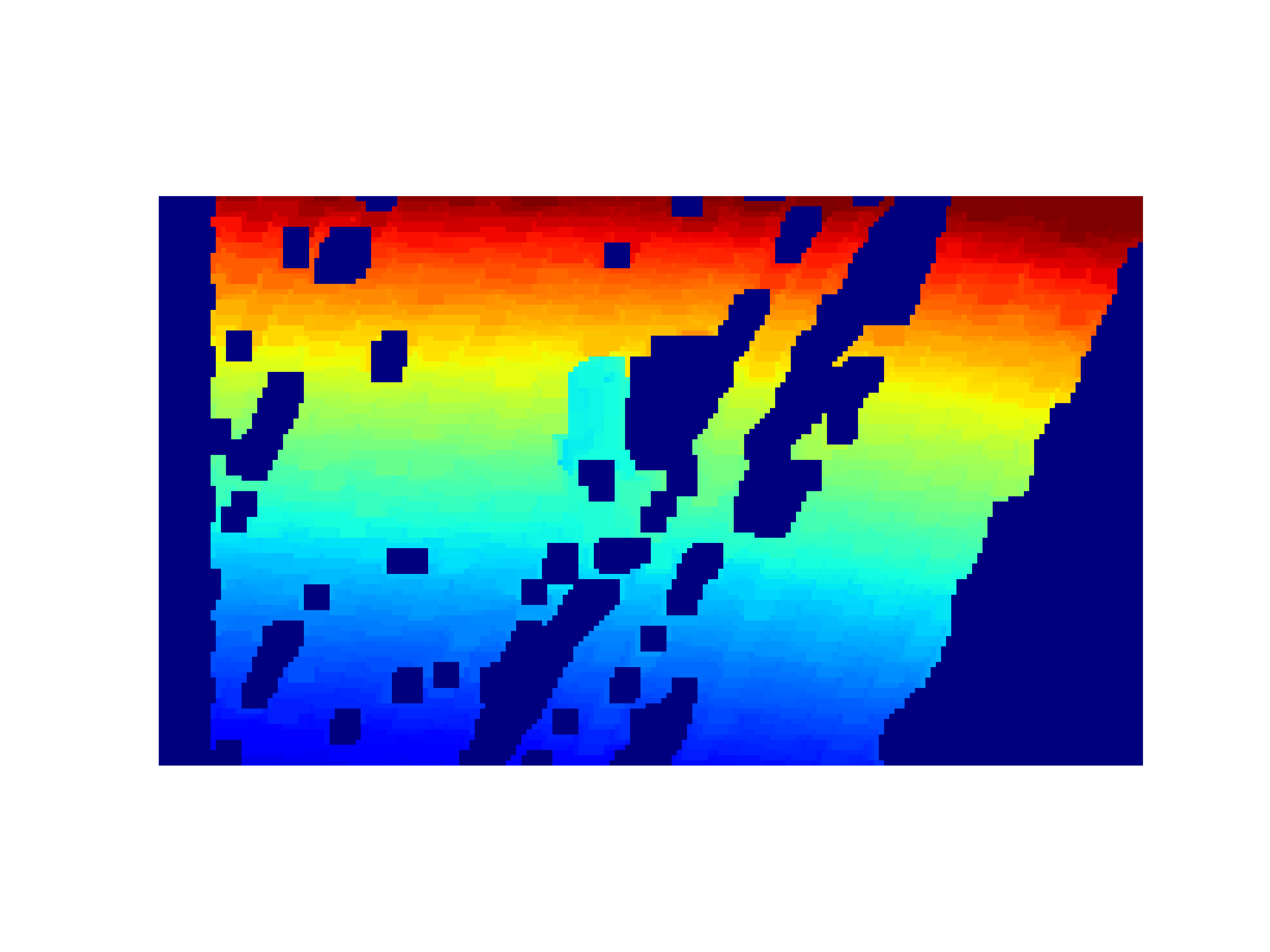}}
		&\frame{\includegraphics[trim={18cm, 5.0cm, 19cm, 5.0cm},clip,width=\linewidth]{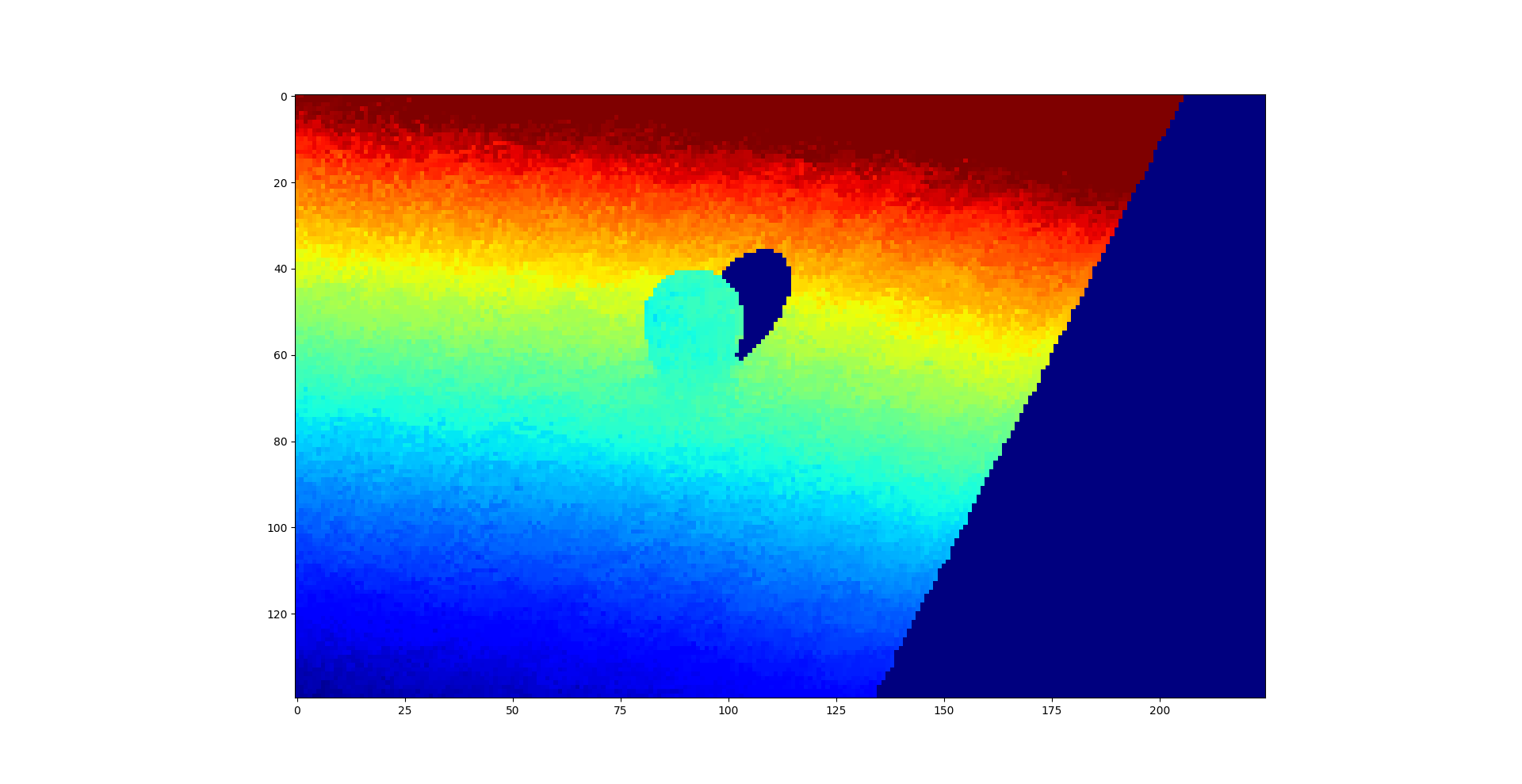}}
		&\frame{\includegraphics[trim={6cm, 2.5cm, 5cm, 2.9cm},clip,width=\linewidth]{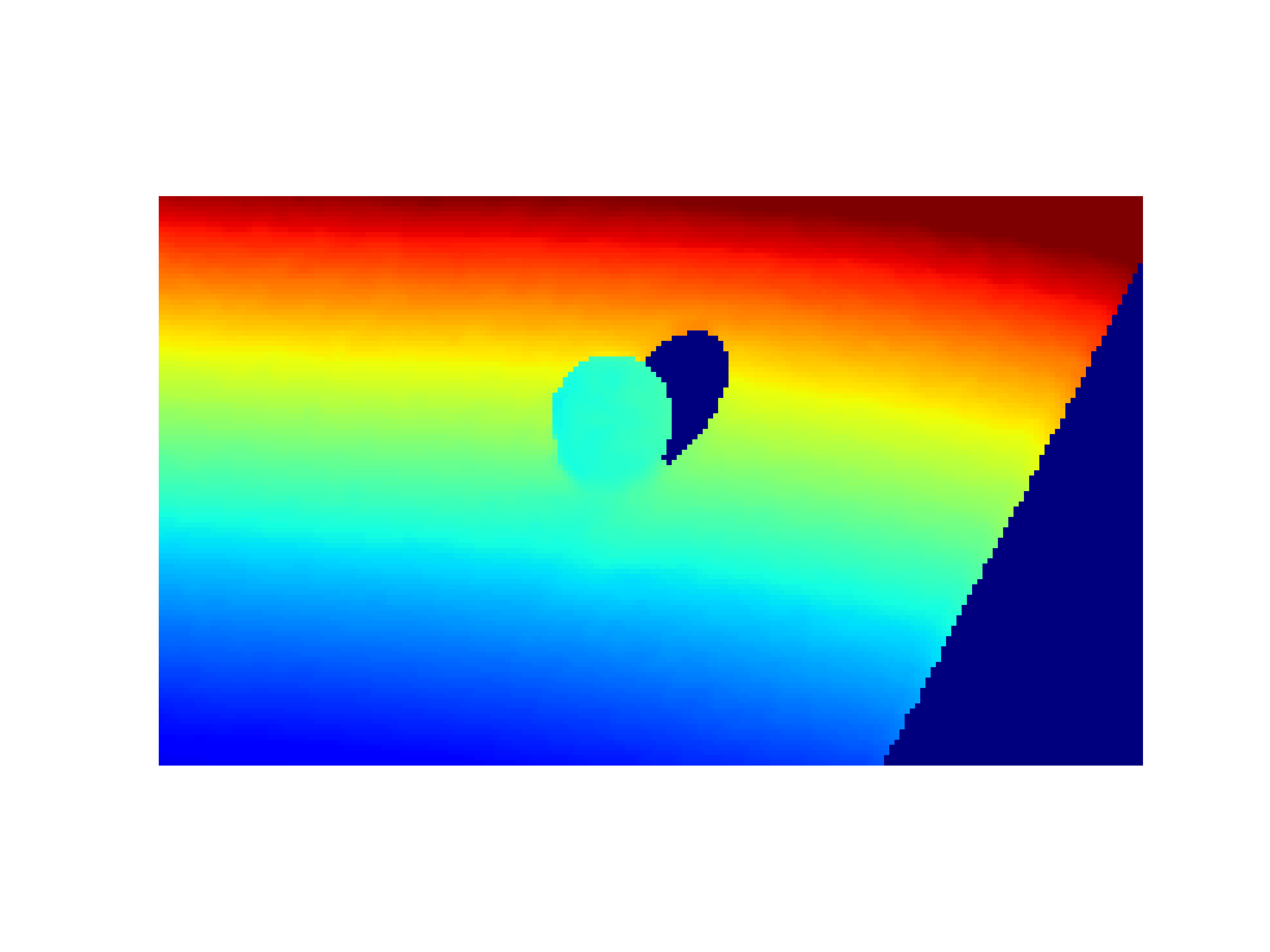}}
		&\frame{\includegraphics[trim={15cm, 5cm, 12.5cm, 9cm},clip,width=\linewidth]{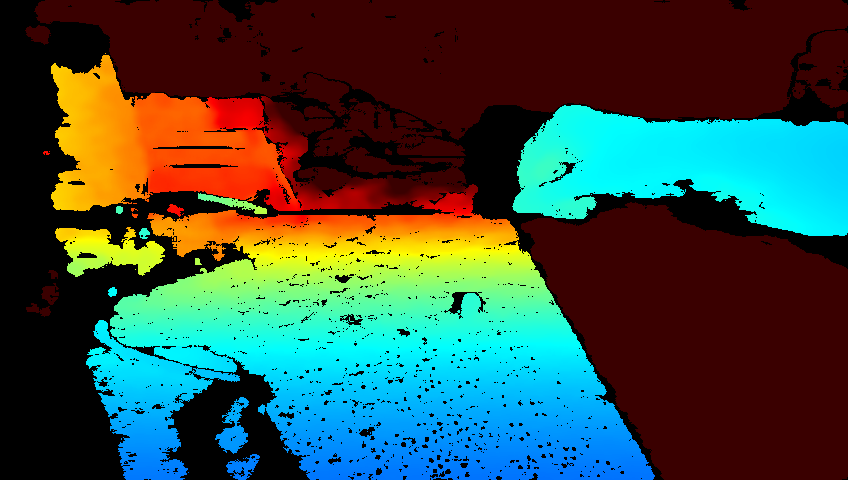}}
		\\
		
	\end{tabular}
    \vspace{-1ex}
	\caption{\emph{Dynamic scenes}.
	Depth maps produced by several SL systems on dynamic scenes (spinning tape, fan, bouncing balls, popping balloons, etc.).
	Our method (4th column) produces the most accurate and detailed depth maps compared to the baselines.
	}
	\label{fig:experim:dynamic}
	\vspace{-2ex}
\end{figure}

We also test our method on eight dynamic scenes (Fig.~\ref{fig:experim:dynamic}) with diverse challenging scenarios to show the capabilities of the proposed method to recover depth information in high-speed applications.
Specifically, Fig.~\ref{fig:experim:dynamic} shows depth recovered using our method and the baselines for the eight sequences.
The figure shows a good performance of our technique in fast motion scenes and in the presence of \mbox{(self-)}occlusions (e.g., Scotch tape and Multi-object) and thin structures (e.g., fan).
Objects do not need to be convex to recover depth with the proposed SL system.
We observe that MC3D depth estimation is inaccurate due to inherent noise in the event timestamps.
In the case of tape spin and fan scenes, MC3D depth has significant holes which cannot be recovered even after post-processing.
SGM performs better than MC3D, however its performance decreases in the presence of noise: 
in the origami fan scene, the depth along the wing of the fan and the wall has significant artefacts.
Our method is robust to these artefacts and can accurately estimate depth in challenging scenes.
Qualitative comparison against Intel RealSense shows favorable performance of our event-based SL method compared to frame-based SL for dynamic scenes.

Because it is \emph{challenging} (if not impossible) to obtain \emph{accurate} ground truth depth at \SI{}{\milli\second} resolution in natural dynamic scenes, such as the deforming origami fan rotating at variable speed, spinning tape, etc. we do not report quantitative results. 
In static scenes, we acquire accurate ground truth depth by time-averaging \SI{1}{\second} of scan data. 
However, this is not possible in dynamic scenes. 
The static scene experiments allow us to assess the accuracy of our method (which only requires \SI{16}{\milli\second} of data) and provide a ballpark for the accuracy of dynamic scenes. %

\textbf{Discussion}. 
The experiments show that the proposed method produces, with the input data from a single scan pass, accurate depth maps at high frequency.
This was possible by exploiting local event correlations at the expense of increasing the computational effort compared to MC3D.
The current Python implementation of the proposed method is 38 times slower than MC3D.
Nevertheless, we think this can be optimized further for real-time operation.

We also found that the method suffers in the presence of strong specularities (coin sequence, bike scene). 
Still, our method is able to handle specularities better than passive systems that process images using the brightness constancy assumption, which breaks down in these scenarios.

\section{Conclusion}
\label{sec:conclusion}

We have introduced a novel method for depth estimation using a laser point-projector and an event camera. 
The method aims at exploiting correlations between events (sparse space-time measurements), which previous methods on the same setup had not explored.
We formulated the problem from first principles, aiming at maximizing spatio-temporal consistency while formulating the problem 
in an amenable stereo fashion.
The experiments showed that the proposed method outperforms the frame-based (Intel RealSense) and event-based baselines (MC3D, SGM),
producing, given input data from a single scan pass, similar 3D reconstruction results as the temporal average of 60 scans with MC3D.
The method also provides best results in dynamic scenes and under broad illumination conditions. 
Exploiting local correlations was possible by introducing more event processing effort into the system. 
The effect of post-processing on the output of our method was marginal, signaling a thoughtful design.
Finally, we think that the ideas presented here can spark a new set of techniques for high-speed depth acquisition and denoising with event-based structured light systems.

\section*{Acknowledgement}
\up{We thank Dr.~Dario Brescianini and Kira Erb for their help with the prototype and data collection.}
\title{\MYTITLE\\---Supplementary Material---}
\maketitle
\ifshowpagenumbers
\else
\thispagestyle{empty}
\fi

\section*{MC3D Baseline}
We implemented the state-of-the-art method proposed in \cite{Matsuda15iccp}.
Moreover, we improved it by removing the need to scan the two end-planes of the scanning volume, which were used to linearly interpolate depth, as we explain.

The method in \cite{Matsuda15iccp} required to scan two planes at known distances from the setup at the two ends of the scanning volume.
These planes were used for calibration and depth estimation.
If $d_n, d_f$ are the disparities corresponding to these two %
planes at depths $Z_n, Z_f$ (near and far, respectively), 
then the depth $Z$ at a pixel $(x,y)$ with disparity $d(x,y)$ was linearly interpolated by~\cite{Wang21jsen}:
\begin{equation}
Z(x,y) = Z_n + Z_f \, \frac{d(x,y) - d_n(x,y)} {d_f(x,y) - d_n(x,y)}
\end{equation}
This first-order method, which assumes pinhole models and a small illumination angle approximation throughout the scan volume,
was justified in \cite{Matsuda15iccp} to overcome the low spatial resolution of the DVS128 ($128 \times 128$ pixels) and the jitter in the event timestamps.

In contrast to the setup in~\cite{Matsuda15iccp}, we use a higher resolution ($\approx20\times$) event camera and calibrate using events.
Therefore, we can estimate depth from disparity without the need for prior scanning of the end-planes.
In our version of MC3D, depth is given by the classical triangulation equation for a canonical stereo configuration 
(assuming the image planes of the projector and event camera are rectified using the calibration information):
\begin{equation}
\label{eq:depth_mc3d}
Z(\bxc) = b \frac{F}{|\bxc - \bxp|},
\end{equation}
where $b$ is the stereo baseline, $F$ is the focal length, and the denominator is the disparity.

\cleardoublepage
{\small
\bibliographystyle{ieeetr_fullname} %
\bibliography{main.bbl}
}

\end{document}